\documentclass[letterpaper]{article}
\usepackage{aaai2026}
\usepackage{times}
\usepackage{helvet}
\usepackage{courier}
\usepackage[hyphens]{url}
\usepackage{graphicx}
\usepackage{natbib}
\usepackage{caption}
\usepackage{graphicx}
\usepackage[ruled,vlined]{algorithm2e}
\usepackage{multirow}
\usepackage{bbding}
\usepackage{amsmath}
\usepackage{amssymb}
\usepackage{booktabs}
\usepackage{array}
\usepackage{colortbl}
\usepackage{tablefootnote}
\usepackage{pifont}
\usepackage{caption}
\usepackage{subcaption}
\usepackage{enumitem}
\renewcommand{\tablename}{Table}
\renewcommand{\figurename}{Figure}

\newcommand{\etc}{\textit{etc.}}

\definecolor{LightYellow}{rgb}{1.0,1.0,0.84}
\definecolor{LightGreen}{rgb}{0.9,1.0,0.88}
\definecolor{LightCyan}{rgb}{0.9,1,1}
\definecolor{LightBlue}{rgb}{0.9,0.94,1}
\definecolor{LightIndigo}{rgb}{0.92,0.9,1}
\definecolor{LightMagenta}{rgb}{0.95,0.88}
\definecolor{LightGray}{rgb}{0.9,0.9,0.9}

\title{X-SAM: From Segment Anything to Any Segmentation}
\author {
Hao Wang\textsuperscript{\rm 1,2},
    Limeng Qiao\textsuperscript{\rm 3},
    Zequn Jie\textsuperscript{\rm 3},
    Zhijian Huang\textsuperscript{\rm 1},
    Chengjian Feng\textsuperscript{\rm 3},\\
    Qingfang Zheng\textsuperscript{\rm 2},
    Lin Ma\textsuperscript{\rm 3},
    Xiangyuan Lan\textsuperscript{\rm 2 \corrauthor},
    Xiaodan Liang\textsuperscript{\rm 1 \corrauthor}
}
\affiliations {
\textsuperscript{\rm 1}Sun Yat-sen University\\
\textsuperscript{\rm 2}Peng Cheng Laboratory\\
\textsuperscript{\rm 3}Meituan Inc.\\
{\{wanghao9610, xdliang328\}@gmail.com, lanxy@pcl.ac.cn}
}

\usepackage{bibentry}

\begin{document}

\maketitle
\begin{abstract}
    Large Language Models (LLMs) demonstrate strong capabilities in broad knowledge representation, yet they are inherently deficient in pixel-level perceptual understanding. Although the Segment Anything Model (SAM) represents a significant advancement in visual-prompt-driven image segmentation, it exhibits notable limitations in multi-mask prediction and category-specific segmentation tasks, and it cannot integrate all segmentation tasks within a unified model architecture. To address these limitations, we present X-SAM, a streamlined Multimodal Large Language Model (MLLM) framework that extends the segmentation paradigm from \textit{segment anything} to \textit{any segmentation}. Specifically, we introduce a novel unified framework that enables more advanced pixel-level perceptual comprehension for MLLMs. Furthermore, we propose a new segmentation task, termed Visual GrounDed (VGD) segmentation, which segments all instance objects with interactive visual prompts and empowers MLLMs with visual grounded, pixel-wise interpretative capabilities. To enable effective training on diverse data sources, we present a unified training strategy that supports co-training across multiple datasets. Experimental results demonstrate that X-SAM achieves state-of-the-art performance on a wide range of image segmentation benchmarks, highlighting its efficiency for multimodal, pixel-level visual understanding. Code is available at \url{https://github.com/wanghao9610/X-SAM}.
\end{abstract}

\section{Introduction}

\begin{figure}[!htbp]
    \centering
    \includegraphics[width=0.96\linewidth]{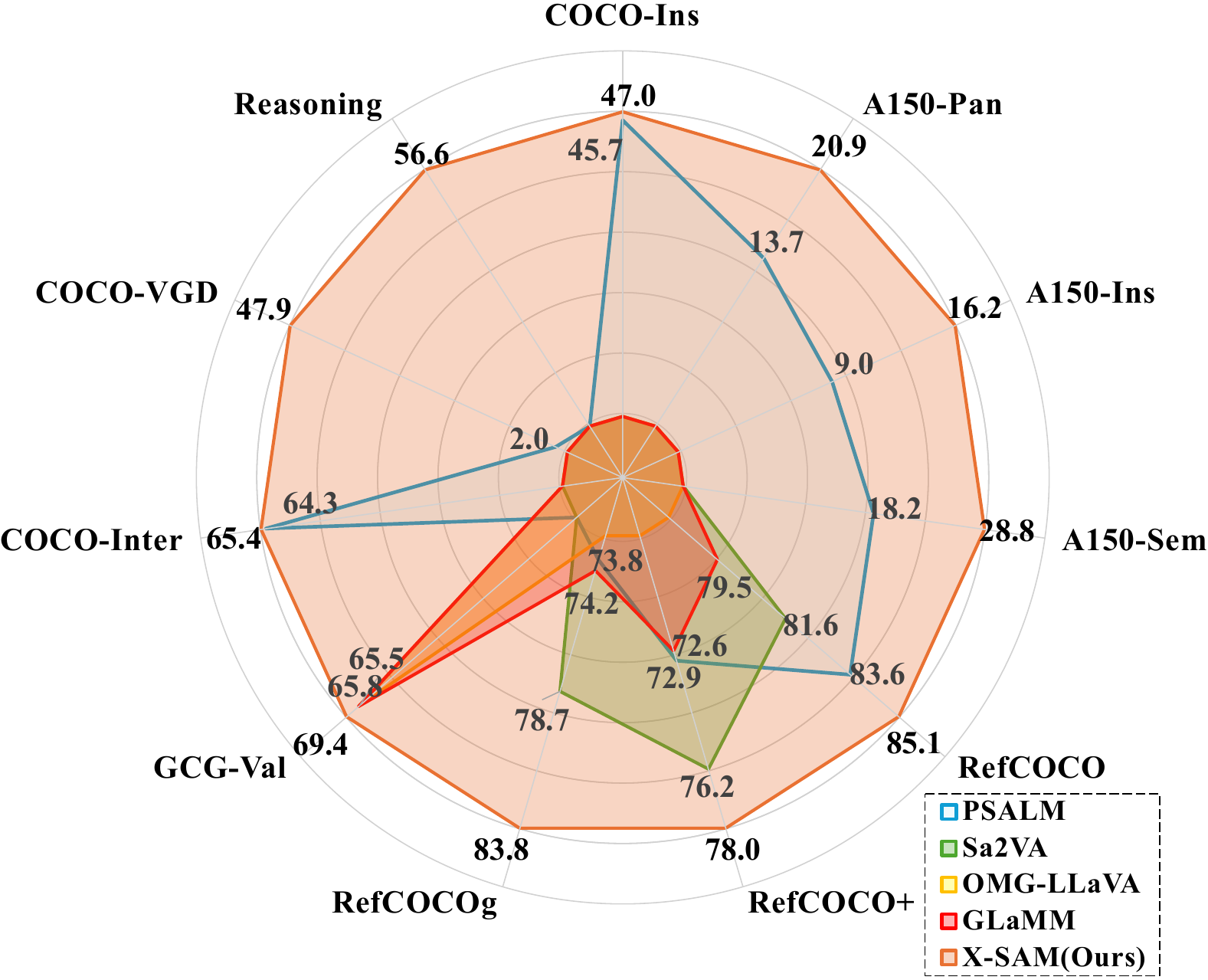}
    \caption{Illustration of Performance of X-SAM on Image Segmentation Benchmarks. X-SAM consistently surpasses existing Multimodal Large Language Models (MLLMs) across all evaluated segmentation benchmarks.}
    \vspace{-0.2cm}
    \label{fig:performance}
\end{figure}

Multi-modal Large Language Models (MLLMs) have exhibited substantial advancements alongside the rapid development of Large Language Models (LLMs)~\cite{bai2023qwen, touvron2023llama} and multi-modal pre-training methods~\cite{radford2021clip, jia2021align}. These models have shown remarkable effectiveness in a wide range of applications, including image captioning~\cite{xu2015show}, VQA~\cite{antol2015vqa}, and visual editing~\cite{chen2018imgedit}. However, current MLLMs lack the capability to generate dense pixel-level outputs for precise spatial understanding. This limitation poses a considerable challenge in directly addressing tasks that require pixel-level comprehension of visual data, such as image segmentation, which is the most critical task in the field of computer vision.

The Segment Anything Model (SAM) represents a foundational segmentation model that demonstrates exceptional efficacy in generating dense segmentation masks and has inspired the development of various segmentation tasks, such as high-quality segmentation~\cite{ke2023hqsam}, matching anything~\cite{li2024matching}, and tracking anything~\cite{rajivc2025track}. Nevertheless, SAM's architecture is fundamentally constrained by its dependency on visual prompts, which significantly limits its direct applicability to a wide range of image segmentation tasks, including generic (semantic, instance, panoptic) segmentation, referring segmentation, and open-vocabulary (OV) segmentation, among others. Achieving a unified framework capable of addressing various image segmentation tasks remains a challenging problem.

\begin{figure*}[!htbp]
    \centering
    \includegraphics[width=1.0\linewidth]{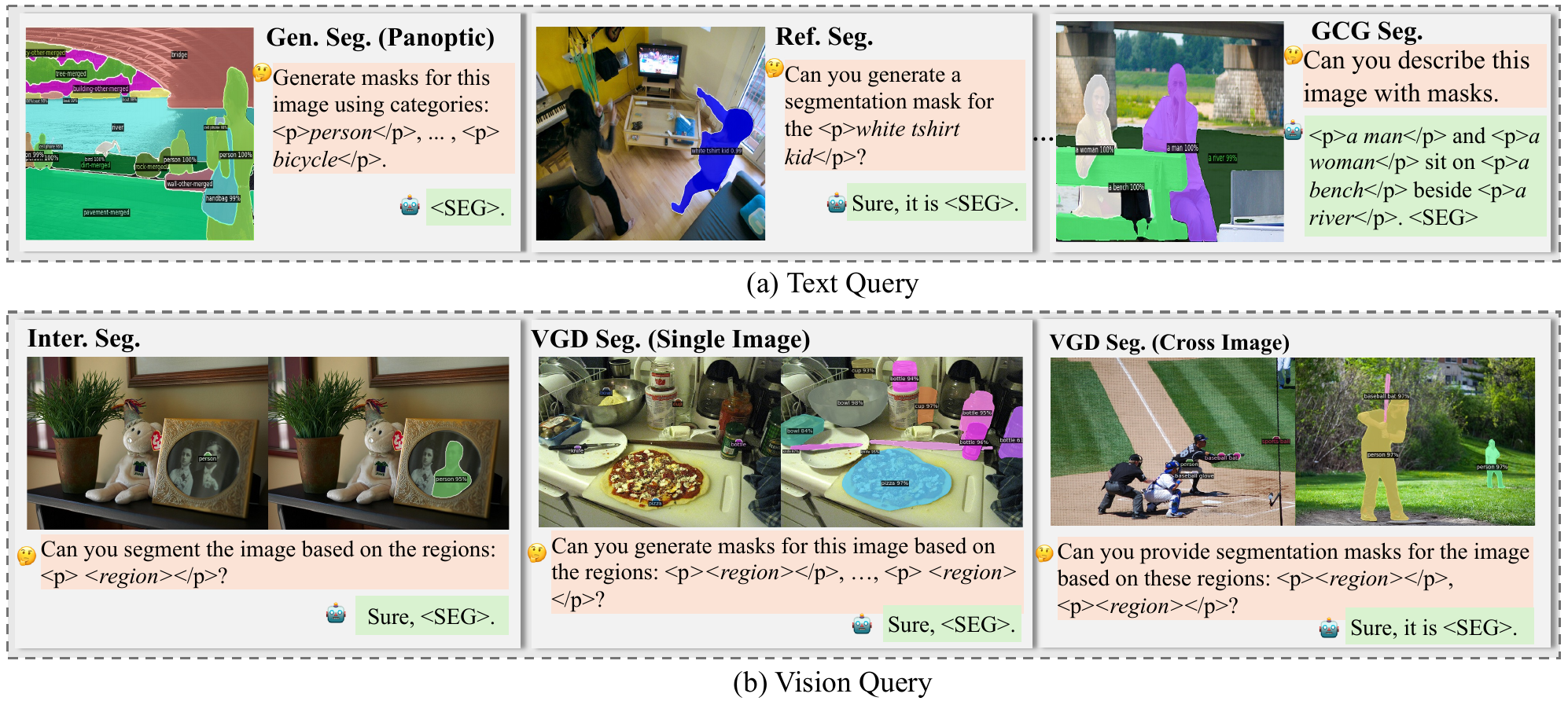}
    \caption{Illustration of the capabilities of X-SAM. (a). Text query tasks: Generic (Gen.), Referring(Ref.), Reasoning(Rea.), and Grounded Conversation Generation(GCG) segmentation, \etc. (b). Vision query tasks: Interactive(Inter.) and Visual GrounDed (VGD) segmentation for single and cross-image.}
    \vspace{-0.2cm}
    \label{fig:capability}
\end{figure*}

In this work, we introduce X-SAM, an innovative framework that unifies diverse image segmentation tasks, expanding the segmentation paradigm from \textit{segment anything} to \textit{any segmentation}. To accomplish this objective, our approach addresses three critical technical challenges: (1) \textit{Task formulation}: Transforming SAM into a versatile segmentation architecture with cross-task applicability. (2) \textit{Modality enhancement}: Augmenting LLMs with multimodal input processing capabilities. (3) \textit{Unified framework}: Developing a cohesive approach to effectively facilitate comprehensive segmentation applications across diverse domains.

First, we develop a unified segmentation MLLM architecture that incorporates a unified mask decoder capable of generating segmentation masks suitable for generalized image segmentation tasks. Second, we expand the multimodal capabilities of MLLMs to process not only textual queries but also visual queries. Specifically, we introduce a novel task termed Visual GrounDed (VGD) segmentation, which segments all instance objects with interactive visual prompts in an image. This task introduces visual guide modalities into large language models (LLMs). Moreover, we propose a unified input format and training methodology that reformulates segmentation tasks within a unified framework, thus optimizing the adaptation of MLLMs to diverse image segmentation tasks.

As shown in \figurename~\ref{fig:capability} and \tablename~\ref{tab:capability}, we present the comprehensive capabilities of X-SAM and compare them with those of other methods. Our proposed framework exhibits capabilities in processing text query-based tasks, such as generic segmentation and referring segmentation, while simultaneously accommodating vision query-based tasks such as interactive segmentation \cite{zhang2024psalm} and our novel VGD segmentation, which functions effectively in both single-image and cross-image contexts. Furthermore, X-SAM leverages the reasoning and generative capacities of LLMs, thereby enabling advanced reasoning segmentation and Grounded Conversation Generation (GCG)~\cite{rasheed2024glamm} segmentation.

X-SAM undergoes co-training with a diverse range of datasets. We perform a comprehensive evaluation on more than twenty segmentation datasets across seven distinct image segmentation tasks, even including the image conversion task. X-SAM achieves the state-of-the-art performance across all image segmentation benchmarks, and establishes a robust new baseline for unified pixel-level image understanding, as illustrated in \figurename~\ref{fig:performance}.
In summary, our contributions are as follows:
\begin{itemize}
    \item We introduce X-SAM, a novel unified framework that extends the segmentation paradigm from \textit{segment anything} to \textit{any segmentation}. Our approach formulates diverse image segmentation tasks into a standardized segmentation format.
    \item We propose a new image segmentation benchmark, Visual GrounDed (VGD) segmentation, which provides visual grounded prompts for MLLMs to segment instance objects in images. The benchmark introduces user-friendly inputs to ground the segmentation objects and guide the MLLMs to output the segmentation masks.
    \item We present a unified multi-stage training strategy to co-train X-SAM with a diverse range of datasets, and conduct extensive evaluations on more than twenty image segmentation benchmarks, achieving state-of-the-art performance on all of them. This establishes a new strong baseline for unified pixel-level perceptual understanding in MLLMs.
\end{itemize}

\section{Related Work}

\begin{table*}[t]
    \centering
    \setlength{\tabcolsep}{2pt}
    \begin{tabular}{l|ccccc|cc}
        \toprule
        \multirow{2}{*}{Method}                             & \multicolumn{5}{c|}{Text Query} & \multicolumn{2}{c}{Vision Query}                                                                                                 \\
        ~                                                   & Gen. Seg.                       & OV Seg.                          & Ref. Seg.             & Rea. Seg      & GCG Seg.      & Inter. Seg.           & VGD Seg.      \\
        \midrule
        \color{gray} SAM\cite{kirillov2023sam}              & \color{gray}~                   & \color{gray}~                    & \color{gray}~         & \color{gray}~ & \color{gray}~ & \color{gray}\ding{51} & \color{gray}~ \\
        \color{gray} Mask2Former\cite{cheng2022mask2former} & \color{gray}\ding{51}           & \color{gray}~                    & \color{gray}~         & \color{gray}~ & \color{gray}~ & \color{gray}~         & \color{gray}~ \\
        \color{gray} ODISE\cite{xu2023odise}                & \color{gray}\ding{51}           & \color{gray}\ding{51}            & \color{gray}~         & \color{gray}~ & \color{gray}~ & \color{gray}~         & \color{gray}~ \\
        \color{gray} UNINEXT\cite{yan2023uninext}           & \color{gray}\ding{51}           & \color{gray}~                    & \color{gray}\ding{51} & \color{gray}~ & \color{gray}~ & \color{gray}\ding{51} & \color{gray}~ \\
        \color{gray} SEEM\cite{zou2023seem}                 & \color{gray}\ding{51}           & \color{gray}\ding{51}            & \color{gray}\ding{51} & \color{gray}~ & \color{gray}~ & \color{gray}\ding{51} & \color{gray}~ \\
        \color{gray} OMG-Seg\cite{li2024omgseg}             & \color{gray}\ding{51}           & \color{gray}\ding{51}            & \color{gray}~         & \color{gray}~ & \color{gray}~ & \color{gray}\ding{51} & \color{gray}~ \\
        \midrule
        LISA\cite{lai2024lisa}                              & ~                               & ~                                & \ding{51}             & \ding{51}     & ~             & ~                     & ~             \\
        GLaMM\cite{rasheed2024glamm}                        & ~                               & ~                                & \ding{51}             & ~             & \ding{51}     & ~                     & ~             \\
        PixelLM\cite{zhang2024psalm}                        & ~                               & ~                                & \ding{51}             & ~             & ~             & ~                     & ~             \\
        OMG-LLaVA\cite{zhang2024omgllava}                   & \ding{51}                       & ~                                & \ding{51}             & ~             & \ding{51}     & ~                     & ~             \\
        Sa2VA\cite{yuan2025sa2va}                           & ~                               & ~                                & \ding{51}             & \ding{51}     & \ding{51}     & ~                     & ~             \\
        PSALM\cite{zhang2024psalm}                          & \ding{51}                       & \ding{51}                        & \ding{51}             & ~             & ~             & \ding{51}             & \ding{51}     \\
        \rowcolor{LightBlue} X-SAM (Ours)                   & \ding{51}                       & \ding{51}                        & \ding{51}             & \ding{51}     & \ding{51}     & \ding{51}             & \ding{51}     \\
        \bottomrule
    \end{tabular}
    \caption{Comparison of Capability. We compare different methods on both segmentation-specific (\textcolor{gray}{Gray}) and MLLM-based.}
    \label{tab:capability}
\end{table*}

\textbf{Multi-modal Large Language Model.}
Multi-modal learning has evolved from early models focused on task-specific fusion and feature extraction~\cite{li2022blip}, to leveraging large language models~\cite{brown2020language, touvron2023llama} for generalized, instruction-tuned multi-task benchmarks~\cite{liu2024mmbench, hudson2019gqa}. LLaVA~\cite{liu2023visual,liu2024improved} introduced visual feature tokenization, inspiring advances in visual representation~\cite{yuan2024osprey}, specialized vision extensions~\cite{lai2024lisa,lin2023video}, and language-guided segmentation~\cite{li2024omgseg,zhang2024omg}. However, most progress remains task-specific. To our knowledge, we are the first to successfully implement a comprehensive approach, opening new directions for image segmentation.

\noindent\textbf{Multi-modal Grounded Segmentation.}
Recent works~\cite{pan2024tokenize,wang2023seggpt} explore visual initiation methods in vision, including learnable tokens~\cite{zhou2022learning}, mask-visual-modeling~\cite{fang2023explore}, and visual prompting encoders~\cite{yuan2024open}. SAM~\cite{kirillov2023segment} and its extensions~\cite{xu2024rap} introduce visual grounding signals to segmentation models, greatly improving performance. Interactive segmentation~\cite{li2024omgseg} further enhances user-guided segmentation for MLLMs. However, existing methods cannot freely treat grounded input as textual input for segmentation. To address this, we propose Visual GrounDed (VGD) segmentation, enabling more diverse multi-modal grounded segmentation.

\noindent\textbf{Unified Segmentation Model.}
Vision transformers~\cite{dosovitskiy2020image} have advanced universal segmentation, with recent works~\cite{xu2024rap,cheng2022masked} developing end-to-end mask classification frameworks that outperform earlier models~\cite{zhou2022transvod,wang2021tmanet} across various applications. Research has expanded to open-world and open-vocabulary segmentation~\cite{yuan2024open,qi2022high,qi2022open}, as well as unified architectures for multiple tasks~\cite{athar2023tarvis,jain2023oneformer,li2024omgseg}. However, most methods focus solely on visual segmentation and lack interactive textual and visual prompts found in MLLMs. To address this, we combine SAM with MLLMs, extending SAM from \textit{segment anything} to \textit{any segmentation}, and introduce a unified framework adaptable to all image segmentation tasks, establishing a strong baseline.

\section{Method}

\begin{figure*}[!htp]
    \centering
    \includegraphics[width=0.95\linewidth]{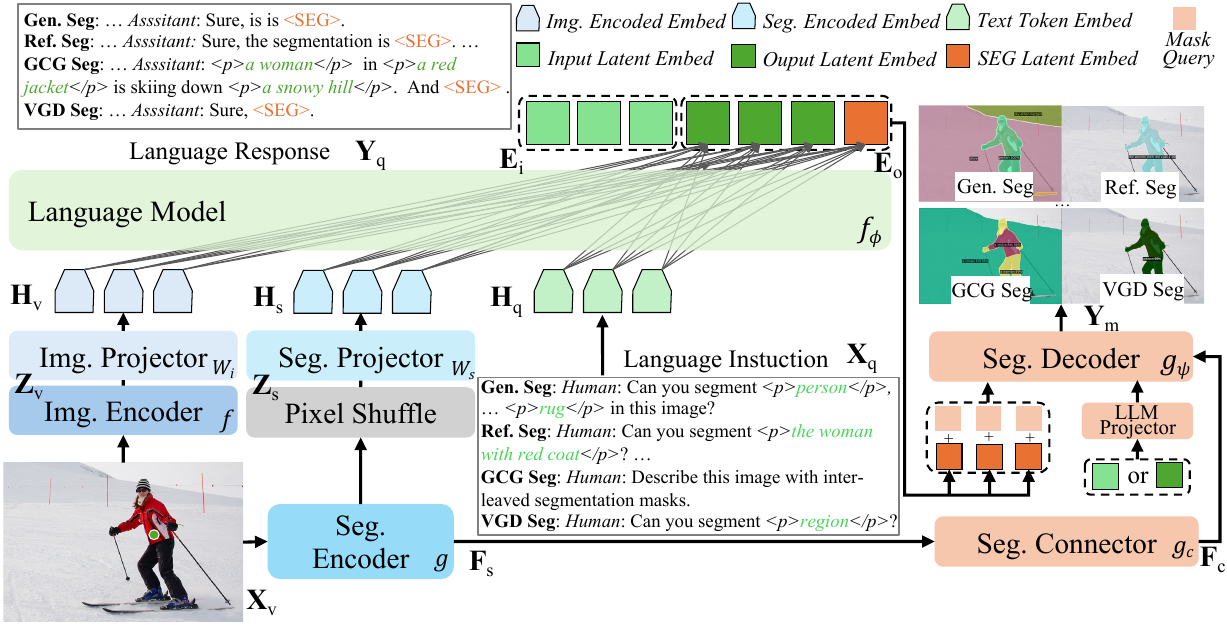}
    \caption{The Overview of X-SAM. X-SAM comprises dual encoders, dual projectors, a language model, a segmentation connector, and a segmentation decoder. The dual encoders process the image and project features to match text embedding dimensions, which are then input to the language model with tokenized text for instruction-guided understanding. The SAM features are connected to the segmentation decoder, which uses the LLM's \texttt{<SEG>} token to generate segmentation masks.}
    \label{fig:framework}
\end{figure*}

To achieve unified image segmentation, we present X-SAM, a novel multi-modal segmentation MLLM. We design a versatile input format and a unified framework to integrate diverse segmentation tasks into a single model. Additionally, we introduce an innovative training strategy that enables SAM to handle any segmentation task. The following sections detail our methodology.

\subsection{Formulation}
The development of a unified segmentation model is fraught with challenges stemming from the diverse nature of segmentation tasks and the variability in input format. To address these issues, we introduce a versatile input format tailored to support a wide range of image segmentation tasks, laying the groundwork for the unified framework of X-SAM. We delineate the input format into two primary categories: text query input and vision query input. The text query input consists exclusively of linguistic prompts derived from user requests, the vision query input integrates both linguistic prompts and visual prompts provided by the user.

\noindent\textbf{Text Query Input.} The majority of existing image segmentation tasks can be conceptualized as text query inputs, including generic segmentation~\cite{kirillov2019panoptic}, referring segmentation, open-vocabulary (OV) segmentation~\cite{li2022ovseg}, GCG segmentation~\cite{rasheed2024glamm}, and reasoning segmentation~\cite{lai2024lisa}. A text query input encapsulates the user's request along with the specific category or object to be segmented, which may be embedded within the user's prompt or generated by a large language model (LLM). To facilitate the GCG segmentation task, inspired by GLaMM~\cite{rasheed2024glamm}, we incorporate two special phrase tokens, \texttt{<p>} and \texttt{</p>}, into the tokenizer to denote the beginning and end of a phrase, respectively. For each category in generic segmentation and GCG segmentation, phrase in referring segmentation, or sentence in reasoning segmentation, the format is standardized as ``\texttt{<p>}category/phrase/sentence\texttt{</p>}''. Specifically, the \texttt{<p>} and \texttt{</p>} tokens are not only encoded in the input tokens but also generated in the output tokens, ensuring consistency across different tasks. Additionally, for the output, we introduce a special token \texttt{<SEG>} into the tokenizer to signify the segmentation result, following the approach in~\cite{lai2024lisa}.

\noindent\textbf{Vision Query Input.} Beyond text query inputs, some tasks necessitate vision query inputs, such as interactive segmentation~\cite{zhang2024psalm} and the Visual GrounDed segmentation proposed in this work. In contrast to text query inputs, vision query inputs incorporate a visual prompt from the user, which may take the form of points, scribbles, boxes, or masks. To denote the visual prompt, we employ a dedicated token, \texttt{<region>}, within the input format. Analogous to the text query input, the visual prompt is formatted as ``\texttt{<p><region></p>}'' and the segmentation output is similarly indicated by the \texttt{<SEG>} token. The \texttt{<region>} token serves as a placeholder for the visual prompt and will be replaced by the region feature extracted from the segmentation encoder.

\noindent\textbf{Unified Formulation.} The latent language embeddings between the \texttt{<p>} and \texttt{</p>} tokens are used as the condition embeddings for the segmentation decoder to compute the classification scores. Based on this formulation, we achieve a unified framework for all image segmentation tasks. Given an input image $\mathbf{X}_\mathrm{v} \in \mathbb{R}^{H \times W \times 3}$ and a language instruction $\mathbf{X}_\mathrm{q} \in \mathbb{R}^{P \times 1}$, the model takes the image and language instruction as inputs and outputs a language response $\mathbf{Y}_\mathrm{q} \in \mathbb{R}^{L \times 1}$ and a segmentation mask $\mathbf{Y}_\mathrm{m} \in \mathbb{R}^{H \times W}$. Here, $P$ is the length of the input text tokens, and $L$ is the total length of the input and output text tokens. $H$ and $W$ denote the height and width of the image, respectively. Detailed input format examples can be found in \figurename~\ref{fig:capability} (a) and (b).

\begin{figure}[!tp]
    \centering
    \includegraphics[width=\linewidth]{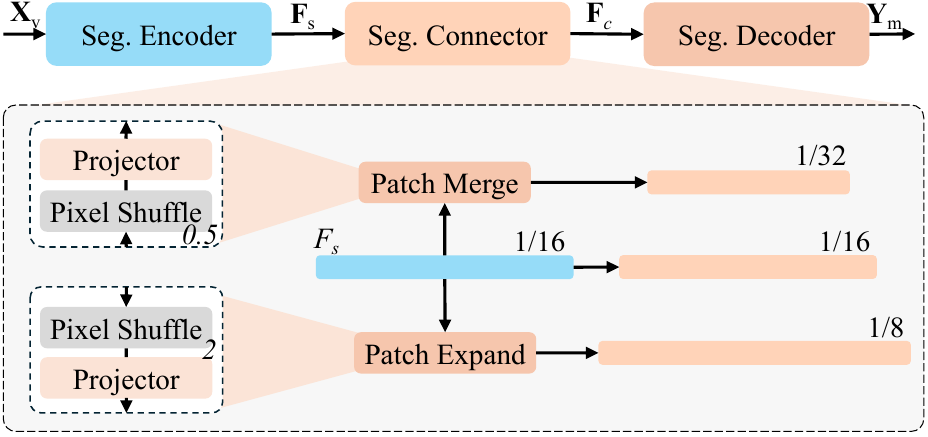}
    \caption{The Architecture of Segmentation Connector.}
    \label{fig:connector}
\end{figure}

\subsection{Architecture}
In this section, we propose X-SAM, a unified segmentation MLLM for any segmentation. As shown in \figurename~\ref{fig:framework}, it includes dual encoders, dual projectors, an LLM, a segmentation connector, and a segmentation decoder.

\begin{table*}[!tp]
    \centering
    \resizebox{\linewidth}{!}{
        \setlength{\tabcolsep}{1.0mm}{
            \begin{tabular}{l|ccccc|cc}
                \toprule
                \multirow{2}{*}{Method}                               & Gen. Seg.                                           & OV Seg.                                               & Ref. Seg.                                     & Rea. Seg.                           & GCG Seg.                       & Inter. Seg.                         & VGD Seg.                          \\
                ~                                                     & Pan.~/~Ins.~/~Sem.                                  & Pan.~/~Ins.~/~Sem.                                    & RefCOCO~/~+~/~g                               & Val~/~Test                          & Val~/~Test                     & Point~/~Box                         & Point~/~Box                       \\
                \midrule
                \color{gray}SAM-L\cite{kirillov2023sam}               & \color{gray}\ding{55}                               & \color{gray}\ding{55}                                 & \color{gray}\ding{55}                         & \color{gray}\ding{55}               & \color{gray}\ding{55}          & \color{gray}51.8~/~76.6             & \color{gray}12.8~/~31.7           \\
                \color{gray} Mask2Former-L\cite{cheng2022mask2former} & \color{gray}57.8~/~48.6~/~67.4                      & \color{gray}\ding{55}                                 & \color{gray}\ding{55}                         & \color{gray}\ding{55}               & \color{gray}\ding{55}          & \color{gray}\ding{55}               & \color{gray}\ding{55}             \\
                \color{gray} SEEM-B\cite{zou2023seem}                 & \color{gray}56.1~/~46.4~/~66.3                      & \color{gray}\ding{55}                                 & \color{gray}-~/~-~/~65.6                      & \color{gray}\ding{55}               & \color{gray}\ding{55}          & \color{gray}47.8~/~44.9             & \color{gray}\ding{55}             \\
                \color{gray} ODISE\cite{xu2023odise}                  & \color{gray}55.4~/~46.0~/~65.2                      & \color{gray}22.6~/~14.4~/~29.9                        & \color{gray}\ding{55}                         & \color{gray}\ding{55}               & \color{gray}\ding{55}          & \color{gray}\ding{55}               & \color{gray}\ding{55}             \\
                \color{gray} OMG-Seg\cite{li2024omgseg}               & \color{gray}53.8~/~-~/~-                            & \color{gray}\ding{55}                                 & \color{gray}\ding{55}                         & \color{gray}\ding{55}               & \color{gray}\ding{55}          & \color{gray}-                       & \color{gray}\ding{55}             \\
                \midrule
                LISA-7B\cite{lai2024lisa}                             & \ding{55}                                           & \ding{55}                                             & 74.9~/~65.1~/~67.9                            & \underline{52.9}~/~\underline{47.3} & \ding{55}                      & \ding{55}                           & \ding{55}                         \\
                GLaMM\cite{rasheed2024glamm}                          & \ding{55}                                           & \ding{55}                                             & 79.5~/~72.6~/~74.2                            & \ding{55}                           & \underline{65.8}~/~64.6        & \ding{55}                           & \ding{55}                         \\
                PixelLM-7B\cite{ren2024pixellm}                       & \ding{55}                                           & \ding{55}                                             & 73.0~/~66.3~/~69.3                            & \ding{55}                           & \ding{55}                      & \ding{55}                           & \ding{55}                         \\
                OMG-LLaVA-7B\cite{zhang2024omgllava}                  & 53.8~/~-~/~-                                        & \ding{55}                                             & 78.0~/~69.1~/~72.9                            & \ding{55}                           & 65.5~/~\underline{64.7}        & \ding{55}                           & \ding{55}                         \\
                Sa2VA-8B\cite{yuan2025sa2va}                          & \ding{55}                                           & \ding{55}                                             & 81.6~/~\underline{76.2}~/~\underline{78.7}    & -~/~-                               & -~/~-                          & \ding{55}                           & \ding{55}                         \\
                PSALM\cite{zhang2024psalm}                            & \textbf{55.9}~/~\underline{45.7}~/~\textbf{66.6}    & \underline{13.7}~/~\underline{9.0}~/~\underline{18.2} & \underline{83.6}~/~72.9~/~73.8                & \ding{55}                           & \ding{55}                      & \underline{64.3}~/~\underline{67.3} & \underline{2.0}~/~\underline{3.7} \\
                \rowcolor{LightBlue} X-SAM (Ours)                     & \underline{54.7}~/~\textbf{47.0}~/~\underline{66.5} & \textbf{20.9}~/~\textbf{16.2}~/~\textbf{28.8}         & \textbf{85.1}~/~\textbf{78.0}~/~\textbf{83.8} & ~\textbf{56.6}~/~\textbf{57.8}      & ~\textbf{69.4}~/~\textbf{69.0} & \textbf{65.4}~/~\textbf{70.0}       & \textbf{47.9}~/~\textbf{49.5}     \\
                \bottomrule
            \end{tabular}}}
    \caption{Comprehensive Performance Comparison. We compare X-SAM to segmentation-specific models (\textcolor{gray}{Gray}) and MLLMs. ``\ding{55}'' denotes unsupported tasks. “–” indicates unreported results. X-SAM achieves state-of-the-art performance across all segmentation tasks with a single model. Best results are in \textbf{bold}, second-best are \underline{underlined}.}
    \label{tab:main}
\end{table*}

\noindent\textbf{Dual Encoders.} There are two encoders in X-SAM, an image encoder and a segmentation encoder. The image encoder $f$ is used to extract the global image feature $\mathbf{Z}_\texttt{v} = f(\mathbf{X}_\texttt{v})$, while the segmentation encoder $g$ extracts the fine-grained image feature $\mathbf{Z}_\texttt{s} = g(\mathbf{X}_\texttt{v})$. The feature from the image encoder is global and benefits image understanding tasks, whereas the feature from the segmentation encoder is fine-grained and benefits image segmentation tasks. We adopt SigLIP2-so400m~\cite{tschannen2025siglip2} as the image encoder and SAM-L~\cite{ke2023hqsam} as the segmentation encoder.

\noindent\textbf{Dual Projectors.} To enhance the LLM's understanding of the image, we concatenate the features from the image encoder and the segmentation encoder before passing them to the LLM. Specifically, the feature from the segmentation encoder is too large to be processed directly by the LLM, so we utilize a pixel-shuffle operation to reduce its spatial size. We then project the reduced feature into the language embedding space $\mathbf{H}_\textit{q}$ via an MLP projector $\mathbf{\textit{W}}_\textit{s}$. For the feature from the image encoder, we directly project it into the language embedding space via an MLP projector $\mathbf{\textit{W}}_\textit{i}$, such that $\mathbf{H}_\textit{v} = \mathbf{\textit{W}}_\textit{i} \cdot \mathbf{Z}_\texttt{v}$ and $\mathbf{H}_\textit{s} = \mathbf{\textit{W}}_\textit{s} \cdot \mathbf{Z}_\texttt{s}$. We then concatenate the features from dual projectors and the language embeddings, and input them into the LLM $f_\phi$.

\noindent\textbf{Segmentation Connector.} For image segmentation tasks, fine-grained multi-scale features are crucial for the segmentation decoder to accurately predict segmentation masks. The output of the segmentation encoder in SAM is single-scale (1/16) with reduced spatial resolution. To obtain multi-scale features, we design a segmentation connector $g_\texttt{c}$, to bridge the segmentation encoder and decoder. As shown in \figurename~\ref{fig:connector}, we perform patch-merge using a pixel-shuffle~\cite{chen2024internvl1.5} with a scale of 0.5 to reduce the spatial size of the last feature in the encoder to a smaller scale (1/32). We also perform patch-expand with a pixel-shuffle of scale 2.0 to increase the spatial size of the last feature to a larger scale (1/8), resulting in multi-scale features for the segmentation decoder.

\noindent\textbf{Segmentation Decoder.} The Segment Anything Model (SAM) can segment a single object based on input text or visual prompts, but it fails to segment all objects in a single inference. To segment all objects at once, we replace its original segmentation decoder with a new decoder, following the approach in~\cite{cheng2022mask2former, vs2024possam}. The segmentation decoder $g_\psi$ predicts masks and their category probabilities from either the input latent embedding $\mathbf{E}_\mathrm{i}$ or the output latent embedding $\mathbf{E}_\mathrm{o}$, multi-scale segmentation features $\mathbf{F}_\mathrm{c}$, and a set of mask query tokens plus the \texttt{<SEG>} token embedding, which bridges the LLM output with the segmentation decoder. Notably, we introduce a latent background embedding to represent the ``ignore'' category for all tasks, thereby unifying all image segmentation tasks with one model.

\begin{table*}[!htp]
    \centering
    \setlength{\tabcolsep}{1.0mm}{
        \begin{tabular}{l|c|ccc|ccc|cc}
            \toprule
            \multirow{2}{*}{Method}                      & \multirow{2}{*}{(M)LLM} & \multicolumn{3}{c|}{RefCOCO} & \multicolumn{3}{c|}{RefCOCO+} & \multicolumn{2}{c}{RefCOCOg}                                                                                                \\
                                                         &                         & val                          & testA                         & testB                        & val              & testA            & testB            & val              & test             \\
            \midrule
            \color{gray} SEEM-L~\cite{zou2023seem}       & \color{gray}-           & \color{gray}-                & \color{gray}-                 & \color{gray}-                & \color{gray}-    & \color{gray}-    & \color{gray}-    & \color{gray}65.6 & \color{gray}-    \\
            \color{gray} UNINEXT-L~\cite{yan2023uninext} & \color{gray}-           & \color{gray}80.3             & \color{gray}82.6              & \color{gray}77.8             & \color{gray}70.0 & \color{gray}74.9 & \color{gray}62.6 & \color{gray}73.4 & \color{gray}73.7 \\
            \color{gray} UNINEXT-H~\cite{yan2023uninext} & \color{gray}-           & \color{gray}82.2             & \color{gray}83.4              & \color{gray}81.3             & \color{gray}72.5 & \color{gray}76.4 & \color{gray}66.2 & \color{gray}74.7 & \color{gray}76.4 \\
            \midrule
            GLaMM~\cite{rasheed2024glamm}                & Vicuna-7B               & 79.5                         & 83.2                          & 76.9                         & 72.6             & 78.7             & 64.6             & 74.2             & 74.9             \\
            OMG-LLaVA~\cite{zhang2024omgllava}           & InternLM-7B             & 77.2                         & 79.8                          & 74.1                         & 68.7             & 73.0             & 61.6             & 71.7             & 71.9             \\
            Sa2VA\cite{yuan2025sa2va}                    & InternVL2-8B            & 81.6                         & -                             & -                            & 76.2             & -                & -                & 78.7             & -                \\
            PSALM~\cite{zhang2024psalm}                  & Phi-1.5-1.3B            & 83.6                         & 84.7                          & \underline{81.6}             & 72.9             & 75.5             & 70.1             & 73.8             & 74.4             \\
            HyperSeg~\cite{wei2024hyperseg}              & Mipha-3B                & \underline{84.8}             & \underline{85.7}              & \textbf{83.4}                & \textbf{79.0}    & \textbf{83.5}    & \textbf{75.2}    & \underline{79.4} & \underline{78.9} \\
            \rowcolor{LightBlue} X-SAM (Ours)            & Phi-3-3.8B              & \textbf{85.1}                & \textbf{87.1}                 & \textbf{83.4}                & \underline{78.0} & \underline{81.0} & \underline{74.4} & \textbf{83.8}    & \textbf{83.9}    \\
            \bottomrule
        \end{tabular}}
    \caption{Comparison of Referring Segmentation. We evaluate methods on referring segmentation benchmarks by (M)LLMs.}
    \label{tab:refseg}
\end{table*}

\subsection{Training}
To improve the performance on diverse image segmentation tasks, we propose a novel multi-stage training strategy. The training strategy consists of three stages: segmentor fine-tuning, alignment pre-training, and mixed fine-tuning.

\noindent\textbf{Stage 1: Segmentor Fine-tuning.} As the segmentation decoder is redesigned, we need to train the segmentor to adapt to segment all objects in a single forward pass. We follow the training pipeline in~\cite{cheng2022mask2former}, which trains the model on the popular COCO-Panoptic~\cite{kirillov2019panoptic} dataset. To enable faster convergence during training, we unfreeze all the parameters in the segmentor while training the segmentation encoder with a lower learning rate. The training objective, $\mathcal{L}_{\mathrm{seg}}$, is the same as in~\cite{cheng2022mask2former}, and is defined as the sum of the classification loss $\mathcal{L}_{\mathrm{cls}}$, the mask loss $\mathcal{L}_{\mathrm{mask}}$, and the dice loss $\mathcal{L}_{\mathrm{dice}}$:
\begin{equation}
    \mathcal{L}_{\mathrm{seg}} = \mathcal{L}_{\mathrm{cls}} + \mathcal{L}_{\mathrm{mask}} + \mathcal{L}_{\mathrm{dice}}
\end{equation}
\noindent\textbf{Stage 2: Alignment Pre-training.} To align the language embeddings and visual embeddings, we perform alignment pre-training on the LLaVA-558K dataset, following~\cite{liu2023llava}. We keep the dual encoders and the LLM parameters frozen, and only train the dual projectors. In this way, the image embeddings and segmentation embeddings can be aligned with the pre-trained LLM word embeddings. The training objective for alignment pre-training is an auto-regressive loss $\mathcal{L}_{\mathrm{regressive}}$:
\begin{equation}
    \mathcal{L}_{\mathrm{regressive}} = -\sum_{i=1}^{N} \log p_\theta \left( \mathcal{Y}_q^{[P+i]} | \mathcal{Y}_q^{[:i-1]}, \mathcal{X}_q^{[:i-1]} \right),
\end{equation}
where $\mathcal{X}_q$ is the input sequence $\mathcal{X}_q = [x_1, x_2, ..., x_p] \in \mathbb{R}^{P \times D}$, $\mathcal{Y}_q$ is the output sequence $\mathcal{Y}_q = [y_1, y_2, ..., y_l] \in \mathbb{R}^{L \times D}$, where $L = P + N$ represents the length of output sequence, $D$ represents the hidden size of LLM. $\theta$ is a trainable parameter in LLM, and we only calculate the loss for the generated text.

\noindent\textbf{Stage 3: Mixed Fine-tuning.} X-SAM is co-trained on multiple datasets across diverse tasks in an end-to-end manner. For the image conversation task, we adopt the auto-regressive loss $\mathcal{L}_{\mathrm{regressive}}$ as is common in MLLM training. For the segmentation tasks, we not only use the segmentation loss as in segmentor training, but also add the auto-regressive loss to the training objective. Benefiting from the unified formulation and simple training objective, end-to-end mixed fine-tuning across diverse tasks can be performed within a unified framework. The training objective for mixed fine-tuning can be formulated as:
\begin{equation}
    \mathcal{L}_{\mathrm{total}} =
    \begin{cases}
        \mathcal{L}_{\mathrm{regressive}},                              & \mathrm{conversation} \\
        \mathcal{L}_{\mathrm{regressive}} + \mathcal{L}_{\mathrm{seg}}, & \mathrm{segmentation}
    \end{cases}
\end{equation}

\section{Experiments}
\subsection{Experiment Settings}

\textbf{Datasets and Tasks.}
For segmentor fine-tuning, we train on the COCO-Panoptic~\cite{kirillov2019panoptic} dataset. For alignment pre-training, we utilize the LLaVA-558K~\cite{liu2023llava} dataset. For end-to-end mixed fine-tuning, we incorporate one image conversation dataset and five types of image segmentation datasets into the training process. To balance the training data across these diverse datasets, we set the training epoch to 1 and adjust the resampling rates of different datasets using dataset balance resampling. After training, X-SAM is capable of performing a variety of tasks, including Image Conversation, Generic, Referring, Reasoning, GCG, Interactive, and VGD Segmentation. Additionally, X-SAM supports Open-Vocabulary (OV) (OV-semantic, OV-instance, OV-panoptic) segmentation, enabling it to segment all objects defined by the input prompt, even those never seen before. Note that COCO-VGD is our proposed VGD segmentation dataset, which is built on the COCO2017 dataset.

\noindent\textbf{Evaluation Metrics.}
We conduct extensive experiments to evaluate the performance of X-SAM. For generic segmentation and open-vocabulary segmentation, we use PQ, mIoU, and mAP as the main metrics for panoptic, semantic, and instance segmentation, respectively. For referring segmentation and reasoning segmentation, we adopt cIoU and gIoU as metrics, following~\cite{zhang2024psalm}. For GCG segmentation, we use M, C, AP50, and mIoU as metrics, following~\cite{rasheed2024glamm}. For interactive segmentation, we use mIoU and cIoU, also following~\cite{zhang2024psalm}. For VGD segmentation, we use AP and AP50. For image conversation, we adopt scores from common MLLM benchmarks as the main metrics, following~\cite{liu2023llava}.

\noindent\textbf{Implementation Details.}
We adopt the XTuner~\cite{2023xtuner} codebase for training and evaluation. During segmentor fine-tuning, we train all parameters, set the batch size to 64, and use a learning rate of 1e-5 for the SAM encoder and 1e-4 for the other parameters. The number of training epochs is set to 36. For alignment pre-training, we train only the dual projector parameters, with a batch size of 256, a learning rate of 1e-3, and one training epoch. For end-to-end mixed fine-tuning, we train all parameters, set the batch size to 64, and use a learning rate of 4e-6 for the dual encoders and 4e-5 for the other parameters, with one training epoch. All training is conducted on 16 A100 GPUs. For image conversation evaluation, we use the VLMEvalKit~\cite{duan2024vlmevalkit} codebase to evaluate performance on MLLM benchmarks. For segmentation task evaluation, we follow the settings described in the corresponding papers and repositories.

\begin{table*}[!htp]
    \centering
    \setlength{\tabcolsep}{1.0mm}{
        \begin{tabular}{l|cccc|ccccc}
            \toprule
            \multirow{2}{*}{Methods}                  & \multicolumn{4}{c|}{Val} & \multicolumn{4}{c}{Test}                                                                                                                   \\
                                                      & METEOR                   & CIDEr                    & AP50             & mIoU             & METEOR           & CIDEr            & AP50             & mIoU             \\
            \midrule
            Kosmos-2\cite{peng2023kosmos2}            & \textbf{16.1}            & 27.6                     & 17.1             & 55.6             & \textbf{15.8}    & 27.2             & 17.2             & 56.8             \\
            LISA-7B\cite{lai2024lisa}                 & 13.0                     & 33.9                     & 25.2             & 62.0             & 12.9             & 32.2             & 24.8             & 61.7             \\
            GLaMM-7B$^\dagger$\cite{rasheed2024glamm} & 15.2                     & \underline{43.1}         & 28.9             & \underline{65.8} & 14.6             & 37.9             & 27.2             & 64.6             \\
            OMG-LLaVA-7B\cite{zhang2024omgllava}      & 14.9                     & 41.2                     & \underline{29.9} & 65.5             & 14.5             & \underline{38.5} & \underline{28.6} & \underline{64.7} \\
            \rowcolor{LightBlue} X-SAM (Ours)         & \underline{15.4}         & \textbf{46.3}            & \textbf{33.2}    & \textbf{69.4}    & \underline{15.1} & \textbf{42.7}    & \textbf{32.9}    & \textbf{69.0}    \\
            \bottomrule
        \end{tabular}}
    \caption{Comparison of GCG Segmentation. $\dagger$ indicates pretraining with the GranD dataset~\cite{rasheed2024glamm}.}
    \label{tab:gcgseg}
\end{table*}

\subsection{Main Results}
\begin{table*}[!htp]
    \centering
    \begin{tabular}{l|cc|cc|cc|cc}
        \toprule
        \multirow{2}{*}{Method}                 & \multicolumn{2}{c|}{Point} & \multicolumn{2}{c|}{Scribble} & \multicolumn{2}{c|}{Box} & \multicolumn{2}{c}{Mask}                                                                           \\
                                                & AP                         & AP50                          & AP                       & AP50                     & AP               & AP50             & AP              & AP50            \\
        \midrule
        PSALM$\dagger$~\cite{zhang2024psalm}    & 2.0                        & 3.3                           & \underline{2.8}          & \underline{4.4}          & 3.7              & 5.8              & \underline{2.3} & \underline{3.3} \\
        SAM$\dagger$~\cite{kirillov2023segment} & \underline{12.8}           & \underline{22.8}              & -                        & -                        & \underline{31.7} & \underline{50.1} & -               & -               \\
        \rowcolor{LightBlue} X-SAM (Ours)       & \textbf{47.9}              & \textbf{72.5}                 & \textbf{48.7}            & \textbf{73.4}            & \textbf{49.5}    & \textbf{74.7}    & \textbf{49.7}   & \textbf{74.9}   \\
        \bottomrule
    \end{tabular}
    \caption{Comparison of VGD Segmentation. $\dagger$ indicates evaluation results following X-SAM setting.}
    \label{tab:vgdseg}
\end{table*}

\begin{table}[!htp]
    \centering
    \resizebox{\linewidth}{!}{
        \setlength{\tabcolsep}{0.8mm}{
            \begin{tabular}{c|cccc}
                \toprule
                \multirow{2}{*}{FT} & COCO-Pan              & A150-OV             & RefCOCO             & Reason-Val          \\
                                    & PQ                    & PQ                  & cIoU                & gIoU                \\
                \midrule
                Specific            & 55.3                  & 16.4                & 81.0                & 48.2                \\
                Mixed               & 54.5($\downarrow0.8$) & 22.4($\uparrow6.0$) & 85.4($\uparrow4.4$) & 57.1($\uparrow8.9$) \\
                \bottomrule
            \end{tabular}}}
    \caption{Ablation on Fine-Tuning(FT).}
    \label{tab:mixedft}
\end{table}

\begin{table}[!ht]
    \centering
    \resizebox{\linewidth}{!}{
        \setlength{\tabcolsep}{0.8mm}{
            \begin{tabular}{cc|cccc}
                \toprule
                \multicolumn{2}{c|}{Encoder} & COCO-Pan       & A150-OV             & GCG-Val             & COCO-VGD                                    \\
                Img.                         & Seg.           & PQ                  & PQ                  & mIoU                  & AP                  \\
                \midrule
                ViT                          & -              & 54.5                & 16.4                & 64.8                  & 40.7                \\
                ViT                          & Swin$^\dagger$ & 56.2($\uparrow1.7$) & 18.6($\uparrow2.2$) & 62.5($\downarrow2.3$) & 48.6($\uparrow7.9$) \\
                ViT                          & SAM            & 54.7($\uparrow0.2$) & 20.9($\uparrow4.5$) & 69.4($\uparrow4.6$)   & 47.9($\uparrow7.2$) \\
                \bottomrule
            \end{tabular}}}
    \caption{Ablation on Dual Encoders. Swin$^\dagger$ is initialized from Mask2Former (M2F)~\cite{cheng2022mask2former}.}
    \label{tab:dualenc}
\end{table}
We conduct extensive evaluation on seven segmentation tasks, including Generic, Open-Vocabulary, Referring, Reasoning, GCG, Interactive, and VGD Segmentation.

\noindent\textbf{Overall.} In \tablename~\ref{tab:main}, we compare X-SAM with current segmentation-specific models and MLLMs. X-SAM demonstrates the most comprehensive capabilities. It achieves performance comparable to state-of-the-art in generic segmentation, and achieves the best performance on other benchmarks, with a single model. X-SAM sets a new state-of-the-art record for image segmentation benchmarks. Detailed results for each task are discussed below.

\noindent\textbf{Referring Segmentation.} We evaluate X-SAM on RefCOCO, RefCOCO+, and RefCOCOg, with the results shown in \tablename~\ref{tab:refseg}. X-SAM outperforms PSALM~\cite{zhang2024psalm} by 1.5\% cIoU, 5.1\% cIoU, and 10.0\% cIoU on the validation sets of RefCOCO, RefCOCO+, and RefCOCOg, respectively. Compared to Sa2VA-8B~\cite{yuan2025sa2va}, X-SAM achieves better results with a smaller model size. It shows performance improvements of 3.5\% cIoU, 1.8\% cIoU, and 5.1\% cIoU on RefCOCO, RefCOCO+, and RefCOCOg, respectively.

\noindent\textbf{GCG Segmentation.} Grounded conversation generation demands detailed image and pixel-level understanding, requiring MLLMs to link captioned objects to their segmentation masks. As shown in \tablename~\ref{tab:gcgseg}, X-SAM achieves a significant performance improvement compared to previous methods and obtains the best results on both the \texttt{Val} and \texttt{Test} sets. In terms of image-level understanding, X-SAM outperforms GLaMM~\cite{rasheed2024glamm} by 0.2\% METEOR and 3.2\% CIDEr on the \texttt{Val} set, and by 0.5\% METEOR and 4.8\% CIDEr on the \texttt{Test} set. In terms of pixel-level understanding, X-SAM outperforms OMG-LLaVA~\cite{zhang2024omgllava} by 3.3\% AP and 3.9\% mIoU on the \texttt{Val} set, and by 4.3\% AP and 4.3\% mIoU on the \texttt{Test} set.

\noindent\textbf{VGD Segmentation.} Visual grounded segmentation demands vision query understanding, requiring MLLMs to comprehend the visual modality and segment all related instances. \tablename~\ref{tab:vgdseg} presents the VGD segmentation results. As VGD segmentation is our newly proposed task, we evaluate PSALM~\cite{zhang2024psalm} following X-SAM’s settings. X-SAM outperforms PSALM by 45.9\% AP, 45.9\% AP, 45.8\% AP, and 47.4\% AP on Point, Scribble, Box, and Mask visual prompts, respectively.


\subsection{Abaltions}
We conduct ablation studies on mixed fine-tuning, dual encoders, multi-stage training, and segmentor architecture, presenting selected benchmark results due to space limitations.

\noindent\textbf{Mixed Fine-tuning.} We ablate the impact of mixed fine-tuning on X-SAM’s performance. As shown in \tablename~\ref{tab:mixedft}, mixed fine-tuning improves performance on out-of-domain COCO benchmarks, demonstrating X-SAM's robust segmentation capabilities—for example, a 6.0\% PQ increase on A150-OV and an 8.9\% gIoU increase on Reason-Val. However, it results in a 0.8\% PQ decrease on COCO-Pan due to the challenge of balancing performance in multi-sources training.

\noindent\textbf{Dual Encoders.} We ablate the design of the dual encoders in X-SAM. As shown in \tablename~\ref{tab:dualenc}, dual encoders with either a SAM or Swin encoder benefit VGD segmentation, achieving 7.2\% AP and 7.9\% AP on COCO-VGD, respectively. Moreover, dual encoders with a SAM encoder consistently improve performance on GCG-Val and A150-OV, while the Swin encoder, which lacks robust segmentation capabilities, provides only a small improvement on A150-OV and even has a negative impact on GCG-Val.

\noindent\textbf{Multi-stage Training.} We ablate the impact of the multi-stage training strategy. As shown in \tablename~\ref{tab:mstage}, the S1 segmentor fine-tuning phase boosts the segmentation capability, producing a notable improvement of 9.3\% PQ in COCO-Pan and 1.5\% PQ in the A150-OV datasets. Meanwhile, the S2 alignment pre-training phase enhances image understanding capabilities, contributing an additional 2.1\% Accuracy on Conv.-MMB. By integrating these stages, X-SAM demonstrates robust advances in image segmentation and comprehension, establishing its effectiveness in addressing complex visual tasks.

\noindent\textbf{Segmentor Architecture.} We ablate the impact of segmentor architecture by performing segmentor fine-tuning for 12 epochs. As shown in \tablename~\ref{tab:segmentor}, M2F decoder brings a large improvement with 9.2\% PQ as the effective design of M2F. The convolution connector performs better than the MLP connector, as the convolution spatial-awareness benefits segmentation, and multi-scale further improves the performance(10.7\% PQ) with more diverse scale features.

\begin{table}[!tp]
    \centering
    \resizebox{\linewidth}{!}{
        \setlength{\tabcolsep}{0.8mm}{
            \begin{tabular}{l|cccc}
                \toprule
                \multirow{2}{*}{M-Stage} & COCO-Pan            & A150-OV             & GCG-Val             & Conv.-MMB           \\
                ~                        & PQ                  & PQ                  & mIoU                & Acc.                \\
                \midrule
                S3                       & 45.2                & 19.4                & 60.6                & 67.2                \\
                S1, S3                   & 54.5($\uparrow9.3$) & 20.9($\uparrow1.5$) & 65.4($\uparrow4.8$) & 67.4($\uparrow0.2$) \\
                S1, S2, S3               & 54.7($\uparrow9.5$) & 20.9($\uparrow1.5$) & 69.4($\uparrow8.8$) & 69.3($\uparrow2.1$) \\
                \bottomrule
            \end{tabular}}}
    \caption{Ablation on Multi-Stage(M-Stage) Training. S1: Stage 1, S2: Stage 2, S3: Stage 3, Conv.: conversation.}
    \label{tab:mstage}
\end{table}

\begin{table}[!tp]
    \centering
    \resizebox{\linewidth}{!}{
        \setlength{\tabcolsep}{0.8mm}{
            \begin{tabular}{c|c|c|ccc}
                \toprule
                Conn. & Decoder & M-Scale               & PQ                    & AP                    & mIoU                  \\ \midrule
                -     & SAM     & \color{gray}\ding{55} & 40.9                  & 26.3                  & 49.5                  \\
                MLP   & M2F     & \color{gray}\ding{55} & 50.1($\uparrow$ 9.2)  & 38.9($\uparrow$ 12.6) & 60.2($\uparrow$ 10.7) \\
                Con.  & M2F     & \color{gray}\ding{55} & 50.3($\uparrow$ 9.4)  & 39.1($\uparrow$ 12.8) & 60.6($\uparrow$ 11.1) \\
                Con.  & M2F     & \ding{51}             & 51.6($\uparrow$ 10.7) & 41.5($\uparrow$ 15.2) & 61.6($\uparrow$ 12.1) \\ \bottomrule
            \end{tabular}}}
    \caption{Ablation on Segmentor Architecture. Conn.: connector, M-Scale: multi-scale, Con.: convolution, M2F: Mask2Former.}
    \label{tab:segmentor}
\end{table}
\section{Conclusion}
In this work, we propose X-SAM, a unified segmentation MLLM that extends the segmentation paradigm from \textit{segment anything} to \textit{any segmentation}, integrating all image segmentation tasks into a single model. Our method can process various multimodal inputs in MLLMs, including both text and visual queries. Moreover, to equip MLLMs with visual grounded perception capabilities, we introduce a new segmentation task, Visual GrounDed (VGD) segmentation, further extending the capabilities of the unified segmentation model. We conduct extensive experiments across all image segmentation tasks, and X-SAM achieves state-of-the-art performance on each task with a single model.

\section{Acknowledgments}
This work is supported by National Key Research and Development Program of China (2024YFE0203100), Scientific Research Innovation Capability Support Project for Young Faculty (No.ZYGXQNJSKYCXNLZCXM-I28), National Natural Science Foundation of China (NSFC) under Grants No.62476293,  General Embodied AI Center of Sun Yat-sen University, National Natural Science Foundation of China (62402252) and (62536003) Guangdong High-Level Talent Programme (2024TQ08X283).

\bibliography{aaai2026}

\clearpage
\appendix

\begin{table*}[!ht]
    \centering
    \resizebox{\linewidth}{!}{
        \setlength{\tabcolsep}{1.0mm}{
            \begin{tabular}{c|c|c}
                \toprule
                Task                   & Datasets                                                                       & \# Sample \\
                \midrule
                \multicolumn{3}{c}{\textbf{Stage 1: Segmentor Fine-tuning}}                                                         \\
                Generic Segmentation   & COCO Panoptic~\cite{kirillov2019panoptic} (118K)                               & 118K      \\
                \midrule
                \multicolumn{3}{c}{\textbf{Stage 2: Alignment Pre-training}}                                                        \\
                Image Conversation     & LLaVA ~\cite{liu2023llava} (558k)                                              & 558k      \\
                \midrule
                \multicolumn{3}{c}{\textbf{Stage 3: Mixed Fine-tuning   }}                                                          \\
                Image Conversation     & LLaVA-1.5~\cite{liu2023llava} (665K)                                           & 665K      \\
                Generic Segmentation   & COCO Panoptic~\cite{kirillov2019panoptic} (118K)                               & 118K      \\
                VGD Segmentation       & COCO-VGD (117K)                                                                & 117K      \\
                Referring Segmentation & RefCOCO(17K), RefCOCO+(17K), RefCOCOg(22k)                                     & 301K      \\
                GCG Segmentation       & Grand-f~\cite{rasheed2024glamm} (1K), RefCOCOg (19K), PSG (28K), Flickr (148K) & 195K      \\
                Reasoning Segmentation & LISA ReasonSeg~\cite{lai2024lisa} (0.2K)                                       & 0.2K      \\
                \bottomrule
            \end{tabular}}}
    \caption{The Datasets Specification. The datasets used in multi-stage training. \# Sample is the total number of samples in this task.}
    \label{tab:datasets}
\end{table*}

\begin{table*}[!htp]
    \centering
    \begin{tabular}{c|ccc}
        \toprule
        \multirow{2}{*}{Item} & Stage 1                                              & Stage 2                & Stage 3           \\
        ~                     & Segmentor Fine-tuning                                & Alignment Pre-training & Mixed Fine-tuning \\
        \midrule
        batch size            & 64                                                   & 256                    & 64                \\
        training epochs       & 36                                                   & 1                      & 1                 \\
        lr of dual encoders   & 1e-5                                                 & -                      & 4e-6              \\
        lr of other modules   & 1e-4                                                 & 1e-3                   & 4e-5              \\
        optimizer             & \multicolumn{3}{c}{AdamW~\cite{loshchilov2017adamw}}                                              \\
        optimizer momentum    & \multicolumn{3}{c}{$\beta_1=0.9, \beta_2=0.999$}                                                  \\
        weight decay          & 0.05                                                 & 0                      & 0.05              \\
        warmup ratio          & 0.03                                                 & 0.03                   & 0.03              \\
        clip max norm         & 0.01                                                 & 1                      & 1                 \\
        image augmentation    & random scale[0.1, 2.0]                               & -                      & -                 \\
        \bottomrule
    \end{tabular}
    \caption{The Hyper-parameters in Multi-stage Training of X-SAM.}
    \label{tab:params}
\end{table*}
\begin{figure*}[!htp]
    \centering
    \includegraphics[width=\linewidth]{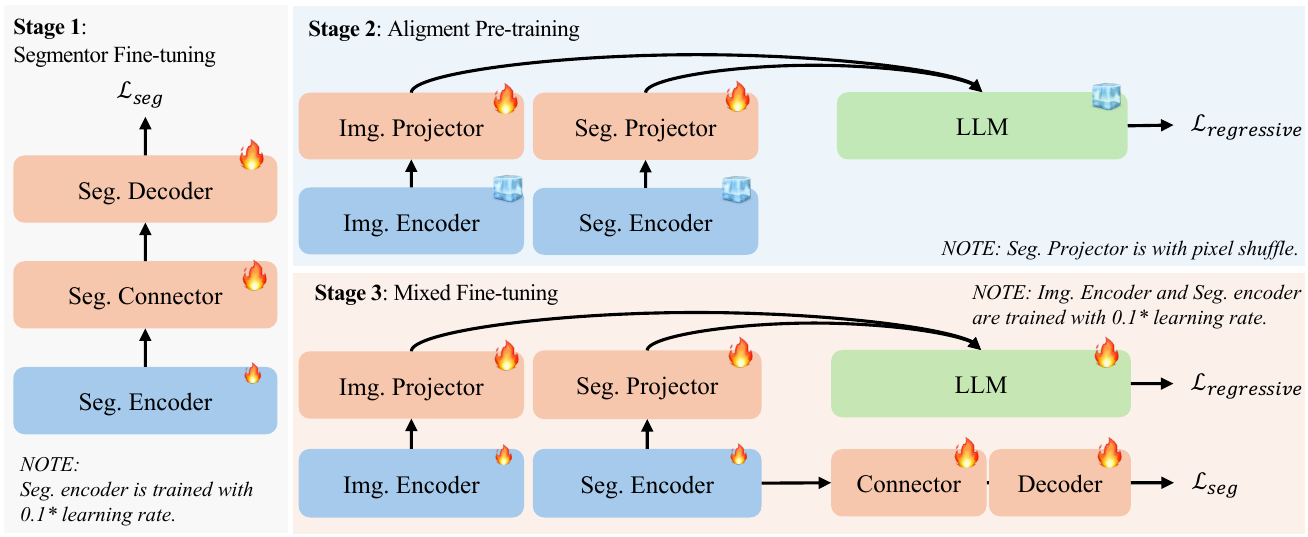}
    \caption{The Multi-stage Training of X-SAM. X-SAM performs a multi-stage training process, including segmentor fine-tuning, alignment pre-training, and mixed fine-tuning. \textit{Segmentor fine-tuning}: train the segmentor on the segmentation datasets to obtain a generalized segmentor. \textit{Alignment pre-training}: train the dual projectors to align the vision features and the LLM features. \textit{Mixed fine-tuning}: fine-tune the dual projectors, the segmentation decoder, and the LLM on the mixed datasets.}
    \label{fig:training}
\end{figure*}

\section{Technical Appendices and Supplementary Material}
In the Appendix, we first provide more details on the dataset, model architecture, and implementation of the proposed method. Then, we present additional experimental results on more benchmarks to demonstrate the effectiveness of our approach. Next, we include ablation studies on dataset balance resampling and the image encoder. Following that, we provide further visualization results for different tasks. Finally, we discuss the limitations and future work.

\subsection{More Dataset Details}
\label{sec:app_datasets}

\textbf{Training Datasets.} In \tablename~\ref{tab:datasets}, we show the datasets used in our multi-stage training. For the segmentor fine-tuning stage, we fine-tune the segmentor on generic segmentation datasets. For the alignment pre-training stage, we pre-train the dual projectors on the alignment LLaVA 558K~\cite{liu2023llava} dataset. For the mixed fine-tuning stage, we fine-tune the whole model on mixed datasets, including both segmentation and conversation datasets. There are six types of datasets in total, including one image-level dataset and five segmentation datasets.

\noindent\textbf{Building COCO-VGD Dataset.} The COCO-VGD dataset is built on the images and annotations of the COCO2017 instance segmentation dataset, which provides instance-level segmentation masks for each object in an image. We automatically generate four types of visual prompts: point, scribe, box, and mask, for each instance in the image, following ~\cite{zhang2024psalm}. We randomly sample one type of visual prompt for each category as the visually grounded prompt during training and evaluation.

\noindent\textbf{Dataset Balance Resampling.} As shown in \tablename~\ref{tab:datasets}, the dataset sizes vary, ranging from 0.2K to 665K. The data ratio among the datasets is crucial to the model’s performance during mixed fine-tuning. To balance the different dataset sizes, we propose a dataset balance resampling strategy following ~\cite{gupta2019lvis}. For each dataset $d$, let $f_d$ be the frequency of dataset $d$ in the mixed datasets. We define the dataset-level repeat factor as $r_d = \max\left(1, \sqrt{t / f_d}\right)$, where $t$ is a hyperparameter controlling the oversampling ratio. We then repeat dataset $d$ with a repeat factor of $r_d$.

\subsection{More Model Details}
\label{sec:app_model}
\textbf{Model Framework.} For the segmentor, we adopt SAM-L~\cite{ke2023hqsam} as the segmentation encoder and the Mask2Former head~\cite{cheng2022mask2former} as the decoder. To reduce the number of connector parameters, we employ a bottleneck architecture, which first reduces the dimension of the segmentation feature to 512 via a 1$\times$1 convolution, then further refines the feature via a 3$\times$3 convolution, and finally expands the dimension to a value determined by the spatial scale of the pixel shuffle~\cite{chen2024internvl1.5} operation using another 1$\times$1 convolution. For the MLLM, we use SigLIP2-so400m~\cite{tschannen2025siglip2} as the image encoder, an MLP as the image projector, another MLP with a pixel shuffle operation as the segmentation projector, and Phi-3-mini-4k-instruct~\cite{abdin2024phi3} as the LLM. The total number of parameters in X-SAM is about 5B.

\noindent\textbf{Region Sampling.} To sample region features from the vision query, we adopt the region sampling strategy from ~\cite{you2023ferret}. Specifically, we first convert the vision query into a binary mask, then perform point sampling on the segmentor-encoded features to obtain the region features, and finally apply mean pooling to produce the final region features. These region features are placed in the corresponding position of \texttt{<region>} in the language instruction, serving as vision query categories for the segmentation decoder.

\subsection{More Training Details}
\label{sec:app_training}
\noindent\textbf{Stage 1: Segmentor Fine-tuning.} During segmentor fine-tuning, we unfreeze all parameters of the segmentor, including the SAM encoder, segmentation connector, and segmentation decoder. The learning rate is set to $\text{1e-4}$ for all components except the segmentation encoder, which uses a learning rate of $\text{1e-5}$. We set the batch size to 64 and train for 36 epochs. The SAM encoder is initialized with pre-trained weights, while the segmentation connector and decoder are initialized with random weights. Additionally, we apply random scale augmentation to the images during training, with a scale of $[0.1, 2.0]$.

\noindent\textbf{Stage 2: Alignment Pre-training.} During alignment pre-training, we train only the parameters of the dual projectors and keep all other parameters fixed. The learning rate is set to $\text{1e-3}$, and the batch size is set to 256. The dual projectors are initialized with random weights, the segmentation encoder is initialized with the pre-trained weights from the segmentor fine-tuning stage, and the image encoder and LLM are initialized with their official pre-trained weights. Training is conducted for 1 epoch.

\noindent\textbf{Stage 3: Mixed Fine-tuning.} During mixed fine-tuning, we fine-tune all parameters of the model. The learning rate for the dual encoders is set to $\text{4e-6}$, while the learning rate for the other modules is set to $\text{4e-5}$. The batch size is set to 64, and training is conducted for 1 epoch. The segmentation encoder, segmentation connector, and segmentation decoder are initialized with pre-trained weights from the segmentor fine-tuning stage, and the image encoder is initialized with official pre-trained weights. The dual projectors are initialized with pre-trained weights from the alignment pre-training stage. Additionally, to make training more stable, we ensure that all data within a global batch come from the same source.

The hyper-parameters in the multi-stage training are shown in \tablename~\ref{tab:params}. A simplified illustration of the multi-stage training is shown in \figurename~\ref{fig:training}.

\begin{table*}[!htp]
    \centering
    \resizebox{\linewidth}{!}{
        \setlength{\tabcolsep}{1.0mm}{
            \begin{minipage}{0.55\linewidth}
                \centering
                \begin{tabular}{l|c|ccc}
                    \toprule
                    Method                                              & Encoder              & PQ               & AP               & mIoU             \\
                    \midrule
                    \color{gray} Mask2Former\cite{cheng2022mask2former} & \color{gray} Swin-L  & \color{gray}57.8 & \color{gray}48.6 & \color{gray}67.4 \\
                    \color{gray} X-Decoder\cite{zou2023xdecoder}        & \color{gray} DaViT-B & \color{gray}56.2 & \color{gray}45.8 & \color{gray}66.0 \\
                    \color{gray} SEEM\cite{zou2023seem}                 & \color{gray} DaViT-B & \color{gray}56.1 & \color{gray}46.4 & \color{gray}66.3 \\
                    \midrule
                    OMG-LLaVA\cite{zhang2024omgllava}                   & ConvNeXt-XXL         & 53.8             & -                & -                \\
                    PSALM\cite{zhang2024psalm}                          & Swin-B               & \textbf{55.9}    & \underline{45.7} & \textbf{66.6}    \\
                    \rowcolor{LightBlue} X-SAM (Ours)                   & SAM-L                & \underline{54.7} & \textbf{47.0}    & \underline{66.5} \\
                    \bottomrule
                \end{tabular}
                \caption{Comparison of Generic Segmentation. We compare different methods on the generic segmentation benchmarks.}
                \label{tab:genseg}
            \end{minipage}\hspace{0.05\linewidth}\begin{minipage}{0.4\linewidth}
                \centering
                \begin{tabular}{l|ccc}
                    \toprule
                    Method                                        & PQ               & AP               & mIoU             \\
                    \midrule
                    \color{gray} MaskCLIP~\cite{ding2022maskclip} & \color{gray}15.1 & \color{gray}6.0  & \color{gray}23.7 \\
                    \color{gray} ODISE\cite{xu2023odise}          & \color{gray}22.6 & \color{gray}14.4 & \color{gray}29.9 \\
                    \midrule
                    PSALM\cite{zhang2024psalm}                    & 13.7             & \underline{9.0}  & 18.2             \\
                    HyperSeg\cite{wei2024hyperseg}                & \underline{16.1} & -                & \underline{22.3} \\
                    \rowcolor{LightBlue} X-SAM (Ours)             & \textbf{20.9}    & \textbf{16.2}    & \textbf{28.8}    \\
                    \bottomrule
                \end{tabular}
                \caption{Comparison of OV Segmentation. We compare different methods on the A150-OV segmentation benchmarks.}
                \label{tab:ovseg}
            \end{minipage}}}
\end{table*}

\begin{table*}[!t]
    \centering
    \begin{tabular}{l|cc|cc|cc|cc}
        \toprule
        \multirow{3}{*}{Method}                  & \multicolumn{2}{c|}{val}     & \multicolumn{6}{c}{test}                                                                                                                                                     \\
        \cline{2-9}
                                                 & \multicolumn{2}{c|}{overall} & \multicolumn{2}{c|}{short query} & \multicolumn{2}{c|}{long query} & \multicolumn{2}{c}{overall}                                                                             \\
        \cline{2-9}
                                                 & gIoU                         & cIoU                             & gIoU                            & cIoU                        & gIoU             & cIoU             & gIoU             & cIoU             \\
        \midrule
        \color{gray} OVSeg~\cite{liang2023ovseg} & \color{gray}28.5             & \color{gray}18.6                 & \color{gray}18.0                & \color{gray}15.5            & \color{gray}28.7 & \color{gray}22.5 & \color{gray}26.1 & \color{gray}20.8 \\
        \color{gray} SEEM~\cite{zou2023seem}     & \color{gray}25.5             & \color{gray}21.2                 & \color{gray}20.1                & \color{gray}11.5            & \color{gray}25.6 & \color{gray}20.8 & \color{gray}24.3 & \color{gray}18.7 \\
        \midrule
        LISA-7B~\cite{lai2024lisa}               & 44.4                         & \underline{46.0}                 & 37.6                            & 34.4                        & 36.6             & 34.7             & 36.8             & 34.1             \\
        LISA-7B~(ft)~\cite{lai2024lisa}          & \underline{52.9}             & \textbf{54.0}                    & \underline{40.6}                & \underline{40.6}            & \underline{49.4} & \textbf{51.0}    & \underline{47.3} & \textbf{48.4}    \\
        \rowcolor{LightBlue} X-SAM (Ours)        & \textbf{56.6}                & 32.9                             & \textbf{47.7}                   & \textbf{48.1}               & \textbf{56.0}    & \underline{40.8} & \textbf{57.8}    & \underline{41.0} \\
        \bottomrule
    \end{tabular}
    \caption{Comparison of Reasoning Segmentation. We compare X-SAM with other methods on the reasoning segmentation benchmark.}
    \label{tab:reaseg}
\end{table*}

\begin{table*}[!tp]
    \centering
    \begin{tabular}{l|cc|cc|cc|cc}
        \toprule
        \multirow{2}{*}{Method}                  & \multicolumn{2}{c|}{Point} & \multicolumn{2}{c|}{Scribble} & \multicolumn{2}{c|}{Box} & \multicolumn{2}{c}{Mask}                                                                             \\
                                                 & mIoU                       & cIoU                          & mIoU                     & cIoU                     & mIoU             & cIoU             & mIoU             & cIoU             \\
        \midrule
        \color{gray} SAM-B\cite{kirillov2023sam} & \color{gray}48.7           & \color{gray}33.6              & \color{gray}-            & \color{gray}-            & \color{gray}73.7 & \color{gray}68.7 & \color{gray}-    & \color{gray}-    \\
        \color{gray} SAM-L\cite{kirillov2023sam} & \color{gray}51.8           & \color{gray}37.7              & \color{gray}-            & \color{gray}-            & \color{gray}76.6 & \color{gray}71.6 & \color{gray}-    & \color{gray}-    \\
        \color{gray} SEEM-B\cite{zou2023seem}    & \color{gray}47.8           & \color{gray}57.8              & \color{gray}43.0         & \color{gray}44.0         & \color{gray}44.9 & \color{gray}42.1 & \color{gray}48.4 & \color{gray}65.0 \\
        \color{gray} OMG-Seg\cite{li2024omgseg}  & \color{gray}59.3           & \color{gray}-                 & \color{gray}-            & \color{gray}-            & \color{gray}-    & \color{gray}-    & \color{gray}-    & \color{gray}-    \\
        \midrule
        PSALM\cite{zhang2024psalm}               & \underline{64.3}           & \textbf{74.0}                 & \textbf{66.9}            & \textbf{80.0}            & \underline{67.3} & \textbf{80.9}    & \underline{67.6} & \textbf{82.4}    \\
        \rowcolor{LightBlue} X-SAM (Ours)        & \textbf{65.4}              & \underline{62.9}              & \textbf{66.9}            & \underline{75.7}         & \textbf{69.6}    & \underline{75.4} & \textbf{69.7}    & \underline{77.0} \\
        \bottomrule
    \end{tabular}
    \caption{Comparison of Interactive Segmentation. We compare X-SAM with other methods on the interactive segmentation benchmark.}
    \label{tab:interseg}
\end{table*}
\subsection{More Evaluation Details}
\label{sec:app_evaluation}
\textbf{PSALM COCO-VGD Evaluation.} PSALM~\cite{zhang2024psalm} is a segmentation MLLM that supports generic segmentation, open-vocabulary segmentation, referring segmentation, interactive segmentation, and more. To evaluate its performance on our proposed COCO-VGD dataset, we follow the same evaluation process as for X-SAM. We randomly sample some instance annotations, as done with X-SAM, and then feed them to PSALM to obtain instance-level predictions. PSALM is trained on the COCO-Interactive dataset, which shares the same source as COCO-VGD but performs classification for each interactive visual prompt. As a result, the AP metric for VGD segmentation is poor because the instance-level predictions are of low quality. This may explain why PSALM lacks the capability for instance-level visual grounding segmentation.

\noindent\textbf{X-SAM COCO-Interactive Evaluation.} X-SAM is the first unified segmentation MLLM, capable of adapting to all image segmentation tasks, including interactive segmentation~\cite{zhang2024psalm}. To evaluate its performance on the COCO-Interactive dataset, we first filter the instance-level predictions using a threshold of 0.5. We then calculate the IoU score between each remaining prediction and the visual prompt mask. Finally, we select the instance prediction with the highest IoU score as the final interactive segmentation result.

\subsection{More Experimental Results}
\label{sec:app_exps}
\begin{table*}[!ht]
    \centering
    \begin{tabular}{l|c|c|cc|cc|cc}
        \toprule
        ~                                         & \multirow{2}{*}{LLM} & \multirow{2}{*}{Zero-Shot} & \multicolumn{2}{c|}{val} & \multicolumn{2}{c|}{testA} & \multicolumn{2}{c}{testB}                                                          \\
        ~                                         & ~                    & ~                          & cIoU                     & gIoU                       & cIoU                      & gIoU             & cIoU             & gIoU             \\
        \midrule
        \color{gray} MattNet~\cite{yu2018mattnet} & \color{gray} -       & \color{gray}\ding{55}      & \color{gray}47.5         & \color{gray}48.2           & \color{gray}58.7          & \color{gray}59.3 & \color{gray}45.3 & \color{gray}46.1 \\
        \color{gray} LTS~\cite{ding2021lts}       & \color{gray}-        & \color{gray}\ding{55}      & \color{gray}52.3         & \color{gray}52.7           & \color{gray}61.9          & \color{gray}62.6 & \color{gray}49.9 & \color{gray}50.4 \\
        \color{gray} CRIS~\cite{wang2022cris}     & \color{gray}-        & \color{gray}\ding{55}      & \color{gray}55.3         & \color{gray}-              & \color{gray}63.8          & \color{gray}-    & \color{gray}51.0 & \color{gray}-    \\
        \midrule
        LISA~\cite{lai2024lisa}                   & Vicuna-7B            & \ding{55}                  & 38.7                     & 32.2                       & 52.6                      & 48.5             & 44.8             & 39.7             \\
        \rowcolor{LightBlue} X-SAM(Ours)          & Phi3-3.8B            & \ding{51}                  & 36.9                     & 44.8                       & 32.3                      & 34.5             & 36.4             & 38.9             \\
        \bottomrule
    \end{tabular}
    \caption{Comparison of Generalized Referring Segmentation. We compare X-SAM with other methods on the gRefCOCO benchmark.}
    \label{tab:grefcoco}
\end{table*}

\begin{table*}[tp]
    \centering
    \setlength{\tabcolsep}{1.0mm}{
        \begin{tabular}{l|lcc|ccc|cc}
            \toprule
            Method                                 & Encoder & \# Params. & Epochs & PQ   & PQ$^\text{Th}$ & PQ$^\text{St}$ & AP$^\text{Th}_\text{pan}$ & mIoU$^\text{pan}$ \\
            \midrule
            Max-DeepLab\cite{wang2021maxdeeplab}   & Max-L   & 451M       & 216    & 51.1 & 57.0           & 42.2           & -                         & -                 \\
            MaskFormer\cite{cheng2022maskformer}   & Swin-L  & 212M       & 300    & 52.7 & 58.5           & 44.0           & 40.1                      & 64.8              \\
            K-Net\cite{zhang2021knet}              & Swin-L  & -          & 36     & 54.6 & 60.2           & 46.0           & -                         & -                 \\
            Mask2Former\cite{cheng2022mask2former} & Swin-L  & 216M       & 100    & 57.8 & 64.2           & 48.1           & 48.6                      & 67.4              \\
            \rowcolor{LightBlue} X-SAM (Ours)      & SAM-L   & 364M       & 36     & 54.2 & 60.0           & 45.4           & 44.2                      & 63.8              \\
            \bottomrule
        \end{tabular}}
    \caption{Comparison of Closed-set Segmentation. We compare X-SAM with other methods on the COCO-Panoptic benchmark.}
    \label{tab:stage1}
\end{table*}

\begin{table*}[!htp]
    \centering
    \begin{tabular}{l|ccccc}
        \toprule
        Method                                       & MME                                & MMBench          & SEED-Bench       & POPE             & AI2D             \\
        \midrule
        \color{gray}LLaVA-1.5~\cite{liu2024improved} & \color{gray}1510~/~-               & \color{gray}64.3 & \color{gray}58.6 & \color{gray}87.3 & \color{gray}-    \\
        \color{gray}LLaVA-OV~\cite{li2024llavaov}    & \color{gray}1580~/~418             & \color{gray}80.8 & \color{gray}75.4 & \color{gray}-    & \color{gray}81.4 \\
        \midrule
        LISA~\cite{lai2024lisa}                      & 1~/~1                              & 0.4              & -                & 0.0              & 0.0              \\
        PixelLM~\cite{ren2024pixellm}                & 309~/~135                          & 17.4             & -                & 0.0              & 0.0              \\
        LaSagnA~\cite{wei2024lasagna}                & 0~/~0                              & 0.0              & -                & 0.0              & 0.0              \\
        GLaMM~\cite{rasheed2024glamm}                & 14~/~9                             & 36.8             & -                & 0.94             & 28.2             \\
        OMG-LLaVA~\cite{zhang2024omgllava}           & \underline{1177}~/~\underline{235} & \underline{47.9} & \underline{56.5} & \underline{80.0} & \underline{42.9} \\
        \rowcolor{LightBlue} X-SAM (Ours)            & \textbf{1374}~/~\textbf{312}       & \textbf{69.3}    & \textbf{69.3}    & \textbf{89.3}    & \textbf{62.6}    \\
        \bottomrule
    \end{tabular}
    \caption{Comparison of Image-level Benchmarks. We compare X-SAM with other methods on the image-level benchmarks, including MME~\cite{fu2024mme}, MMBench~\cite{liu2024mmbench}, SEED-Bench~\cite{li2024seed}, POPE~\cite{li2023pope}, and AI2D~\cite{kembhavi2016ai2d}.}
    \label{tab:vlmeval}
\end{table*}

\noindent\textbf{Generic Segmentation.} \tablename~\ref{tab:genseg} presents the results of generic segmentation. Thanks to our segmentor design and fine-tuning, X-SAM can adapt to generic segmentation and achieves competitive performance on the COCO-Panoptic dataset.

\noindent\textbf{Open-Vocabulary Segmentation.} \tablename~\ref{tab:ovseg} shows the results of open-vocabulary segmentation. Benefiting from the robust mask generation of SAM and our mixed fine-tuning strategy, X-SAM achieves the best performance on open-vocabulary segmentation tasks.

\noindent\textbf{Reasoning Segmentation.} In \tablename~\ref{tab:reaseg}, we present the results of reasoning segmentation on the reasoning segmentation benchmark. We report the performance of our method on both the validation set and the test set, following~\cite{lai2024lisa}. X-SAM achieves the best gIoU metric on both the validation and test sets, even though it is not specifically designed for reasoning segmentation. While the cIoU metric is not the best, it remains comparable to state-of-the-art methods. As the number of samples in the validation and test sets is limited, the results on this benchmark may not be stable.

\noindent\textbf{Generalized Referring Segmentation.} In \tablename~\ref{tab:grefcoco}, we present the results of generalized referring segmentation on the gRefCOCO benchmark. We report the performance of our method on both the validation set and the test set, following~\cite{yu2018mattnet}. X-SAM achieves the comparable performance to state-of-the-art methods under zero-shot setting, it will achieve better performance with the fine-tuning on the gRefCOCO dataset.

\noindent\textbf{Interactive Segmentation.} \tablename~\ref{tab:interseg} shows the results of the interactive segmentation. Since interactive segmentation shares similar data with VGD segmentation, we exclude the training data for interactive segmentation and perform a process similar to that used for VGD segmentation to obtain the interactive segmentation results. X-SAM achieves the best or second-best performance on interactive segmentation, even without being trained on the specific data.

\noindent\textbf{Closed-Set Segmentation.} In \tablename~\ref{tab:stage1}, we present the results of segmentor fine-tuning on the closed-set COCO-Panoptic benchmark. To preserve the robust generalization ability of SAM in mask prediction, we fine-tune the SAM encoder for only 36 epochs. Compared with other methods, our approach achieves comparable performance to the SAM-L encoder with just 36 epochs of fine-tuning without any complex data augmentation.

\noindent\textbf{Image-level Benchmarks.} In \tablename~\ref{tab:vlmeval}, we present the results of image-level benchmarks, including MME~\cite{fu2024mme}, MMBench~\cite{liu2024mmbench}, SEED-Bench~\cite{li2024seed}, POPE~\cite{li2023pope}, and AI2D~\cite{kembhavi2016ai2d}. When jointly co-training with segmentation and conversation datasets, X-SAM achieves the best performance compared to other segmentation MLLMs on these benchmarks. Compared to LISA~\cite{lai2024lisa}, PixelLM~\cite{ren2024pixellm}, and GLaMM~\cite{rasheed2024glamm}, our method achieves significant improvements, demonstrating its effectiveness—even outperforming the previous best, OMG-LLaVA~\cite{zhang2024omgllava}. On the POPE benchmark, X-SAM even surpasses LLaVA-V1.5~\cite{liu2023llava}, which is designed specifically for image-level conversation.

\begin{table*}[!htp]
    \centering
    \begin{tabular}{l|ccccccc|c}
        \toprule
        \multirow{2}{*}{t} & COCO-Pan      & A150-OV          & RefCOCO          & Reason-Val       & GCG-Val          & COCO-VGD         & Conv.-MMB        & \multirow{2}{*}{$\sum\Delta$} \\
        ~                  & PQ            & PQ               & cIoU             & gIoU             & mIoU             & AP               & Acc                                              \\
        \midrule
        0                  & 53.5          & 20.6             & 84.2             & 44.1             & 62.8             & 47.4             & \underline{68.6} & ~~~0.0                        \\
        0.001              & 53.9          & \underline{20.9} & \textbf{85.5}    & 44.1             & 62.7             & 47.7             & 67.3             & $\uparrow0.9$                 \\
        0.01               & 54.3          & \textbf{21.0}    & 84.6             & \underline{46.5} & \underline{63.1} & \textbf{48.1}    & 67.4             & $\uparrow\underline{3.8}$     \\
        0.1                & \textbf{54.7} & \underline{20.9} & \underline{85.1} & \textbf{56.6}    & \textbf{69.4}    & \underline{47.9} & \textbf{69.3}    & $\uparrow\textbf{22.7}$       \\
        \bottomrule
    \end{tabular}
    \caption{Ablation of Dataset Balance Resampling. $\sum\Delta$ represents the total performance improvement compared to the first row.}
    \label{tab:resampling}
\end{table*}

\begin{table*}[!htp]
    \centering
    \resizebox{\linewidth}{!}{
        \setlength{\tabcolsep}{0.8mm}{
            \begin{tabular}{l|cccccc|cc}
                \toprule
                \multirow{2}{*}{Img. Enc.}          & COCO-Pan         & A150-OV          & RefCOCO          & Reason-Val       & GCG-Val          & COCO-VGD         & Conv.-MME                    & Conv.-MMB        \\
                ~                                   & PQ               & PQ               & cIoU             & gIoU             & mIoU             & AP               & Acc                          & Acc              \\
                \midrule
                CLIP~\cite{radford2021clip}         & \underline{53.7} & \underline{21.2} & \underline{84.8} & 55.6             & \underline{62.6} & \textbf{48.0}    & 1327~/~277                   & \underline{68.5} \\
                SigLIP~\cite{zhai2023siglip}        & 53.2             & \textbf{21.7}    & 83.3             & \textbf{61.6}    & 61.8             & \textbf{48.0}    & 1331~/~280                   & 68.4             \\
                SigLIP2~\cite{tschannen2025siglip2} & \textbf{54.7}    & 20.9             & \textbf{85.1}    & \underline{56.6} & \textbf{69.4}    & \underline{47.9} & \textbf{1374}~/~\textbf{312} & \textbf{69.3}    \\ \bottomrule
            \end{tabular}}}
    \caption{Ablation of Image Encoder. We ablate the impact of different image encoders, including CLIP~\cite{radford2021clip}, SigLIP~\cite{zhai2023siglip}, and SigLIP2~\cite{tschannen2025siglip2}, on the performance of X-SAM.}
    \label{tab:encoder}
\end{table*}

\begin{table*}[!htp]
    \centering
    \resizebox{\linewidth}{!}{
        \setlength{\tabcolsep}{0.3mm}{
            \begin{minipage}{0.55\linewidth}
                \centering
                \centering
                \begin{tabular}{l|cccc}
                    \toprule
                    \multirow{2}{*}{LLM}           & COCO-Pan      & COCO-VGD      & Conv.-MMB     \\
                    ~                              & PQ            & AP            & Acc           \\
                    \midrule
                    Phi3-3.8B~\cite{abdin2024phi3} & \textbf{54.7} & \textbf{47.9} & 69.3          \\
                    Qwen3-1.7B~\cite{bai2023qwen}  & 54.3          & 47.8          & 65.8          \\
                    Vicuna-7B~\cite{vicuna2023}    & 54.5          & 47.3          & \textbf{71.3} \\
                    \bottomrule
                \end{tabular}
                \caption{Ablation of LLMs. We ablate the performance of X-SAM with different LLMs.}
                \label{tab:llms}
            \end{minipage}
            \hspace{0.02\linewidth}\begin{minipage}{0.4\linewidth}
                \centering
                \begin{tabular}{c|ccccc}
                    \toprule
                    \multirow{2}{*}{Seg Data} & Conv.-MME                    & SEED          & POPE          & AI2D          \\
                    ~                         & Acc                          & Acc           & Acc           & Acc           \\
                    \midrule
                    \ding{55}                 & \textbf{1397}~/~\textbf{332} & 69.0          & 87.6          & \textbf{64.8} \\
                    \ding{51}                 & 1374~/~312                   & \textbf{69.3} & \textbf{89.3} & 62.6          \\
                    \bottomrule
                \end{tabular}
                \caption{Ablation of Segmentation Data. We ablate the performance of X-SAM with and without segmentation data on the image-level benchmarks.}
                \label{tab:segdata}
            \end{minipage}
        }}
\end{table*}

\subsection{More Ablation Studies}
\label{sec:app_ablations}

\noindent\textbf{Dataset Balance Resampling.} In \tablename~\ref{tab:resampling}, we ablate the oversampling ratio $t$ from 0 to 0.1 in dataset balance resampling. When $t = 0$, no dataset is oversampled. Otherwise, datasets are oversampled with a repeat factor of $r_d = \max\left(1, \sqrt{t / f_d}\right)$. We find that the performance on some small datasets is sensitive to the oversampling ratio $t$, especially for the reasoning segmentation dataset, where the gIoU score improves from 44.1\% to 56.6\% as $t$ increases from 0 to 0.1. Meanwhile, larger datasets are not sensitive to the oversampling ratio $t$. As a result, the overall performance improvement reaches its highest when $t = 0.1$. Therefore, we set $t$ to 0.1 in the final experiment.

\noindent\textbf{Image Encoder.} In \tablename~\ref{tab:encoder}, we ablate the image encoder of X-SAM by replacing it with CLIP~\cite{radford2021clip}, SigLIP-so400m~\cite{zhai2023siglip}, and SigLIP2-so400m~\cite{tschannen2025siglip2}. It can be observed that using more powerful image encoders improves the image content understanding ability of X-SAM, especially on the image conversation and GCG segmentation benchmarks, and even enhances performance on the generic segmentation benchmark. Although SigLIP achieves the best performance on the reasoning segmentation benchmark, it does not have a performance advantage on the other benchmarks. Meanwhile, SigLIP2 demonstrates more robust and consistently better performance across all benchmarks. Therefore, we adopt SigLIP2-so400m as the image encoder for the final experiment.

\noindent\textbf{LLM.} In \tablename~\ref{tab:llms}, we ablate the LLM of X-SAM with different LLMs. We find that Phi3-3.8B~\cite{abdin2024phi3} achieves the best performance on the COCO-Pan and COCO-VGD benchmarks, while Vicuna-7B~\cite{vicuna2023} achieves the best performance on the Conv.-MMB benchmark. Moreover, Phi3-3.8B show more robust and consistent performance across segmentation benchmarks, we adopt it as the LLM for the final experiment.

\noindent\textbf{Segmentation Data.} In \tablename~\ref{tab:segdata}, we ablate the impact of segmentation data on the image-level benchmarks. We find that adding segmentation data improves the performance of X-SAM on the image-level benchmarks, especially on the hallucination-prone benchmarks, such as SEED and POPE.

\subsection{More Visualization Results}
\label{sec:app_vis}
\begin{figure*}[ht]
    \centering
    \captionsetup{skip=5pt,belowskip=-10pt}
    \includegraphics[width=\linewidth]{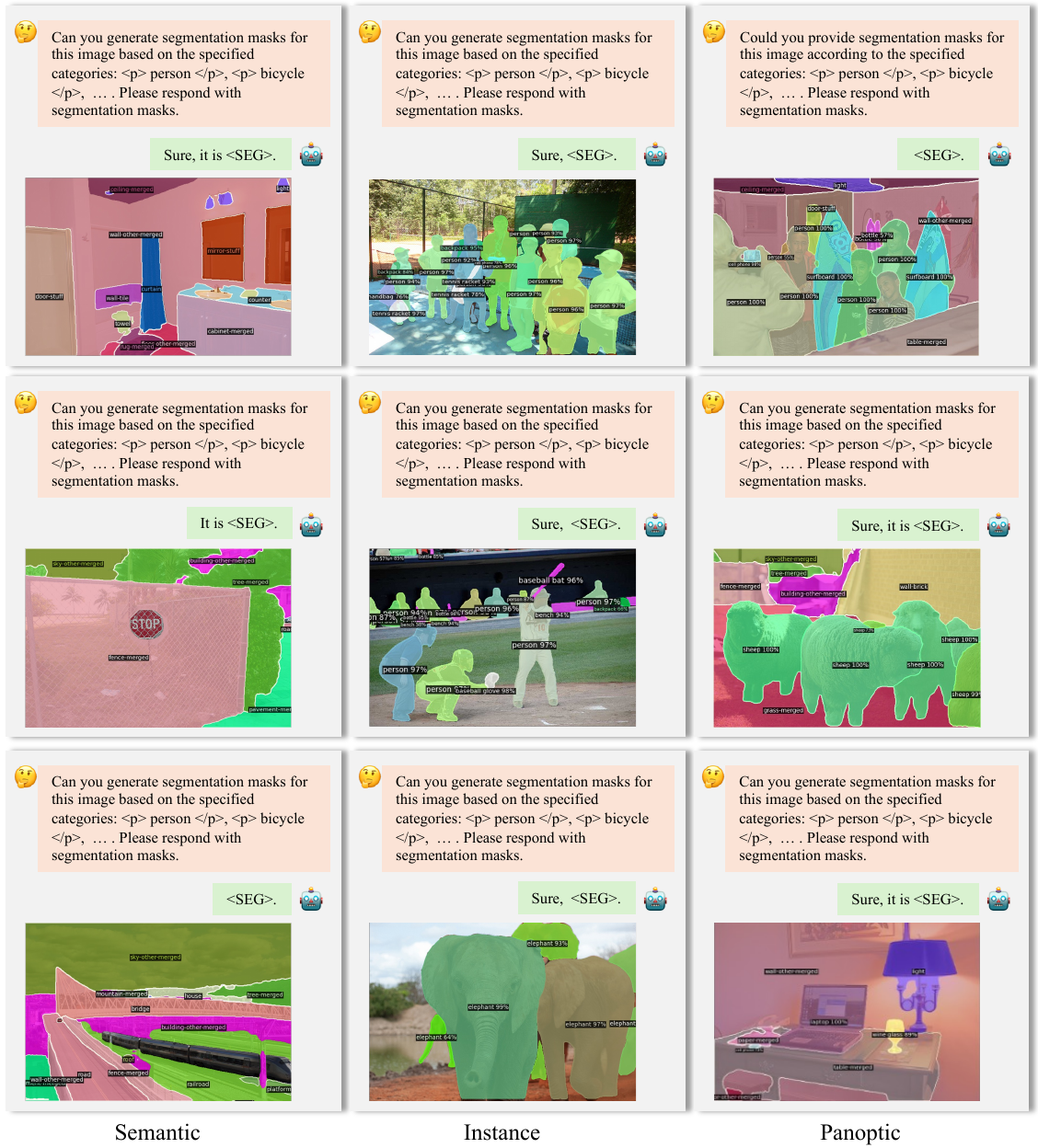}
    \caption{Visualization Results of Generic Segmentation. Visualized images are sampled from the COCO2017 Val set. More category names are omitted for better visualization.}
    \label{fig:genseg_vis}
\end{figure*}

\noindent\textbf{Generic Segmentation.} \figurename~\ref{fig:genseg_vis} shows the visualization results of X-SAM in generic segmentation, including semantic, instance and panoptic segmentation, which needs both the semantic and instance level understanding of the image. X-SAM can generate accurate and complete masks for the objects in the image.

\begin{figure*}[ht]
    \centering
    \includegraphics[width=0.82\linewidth]{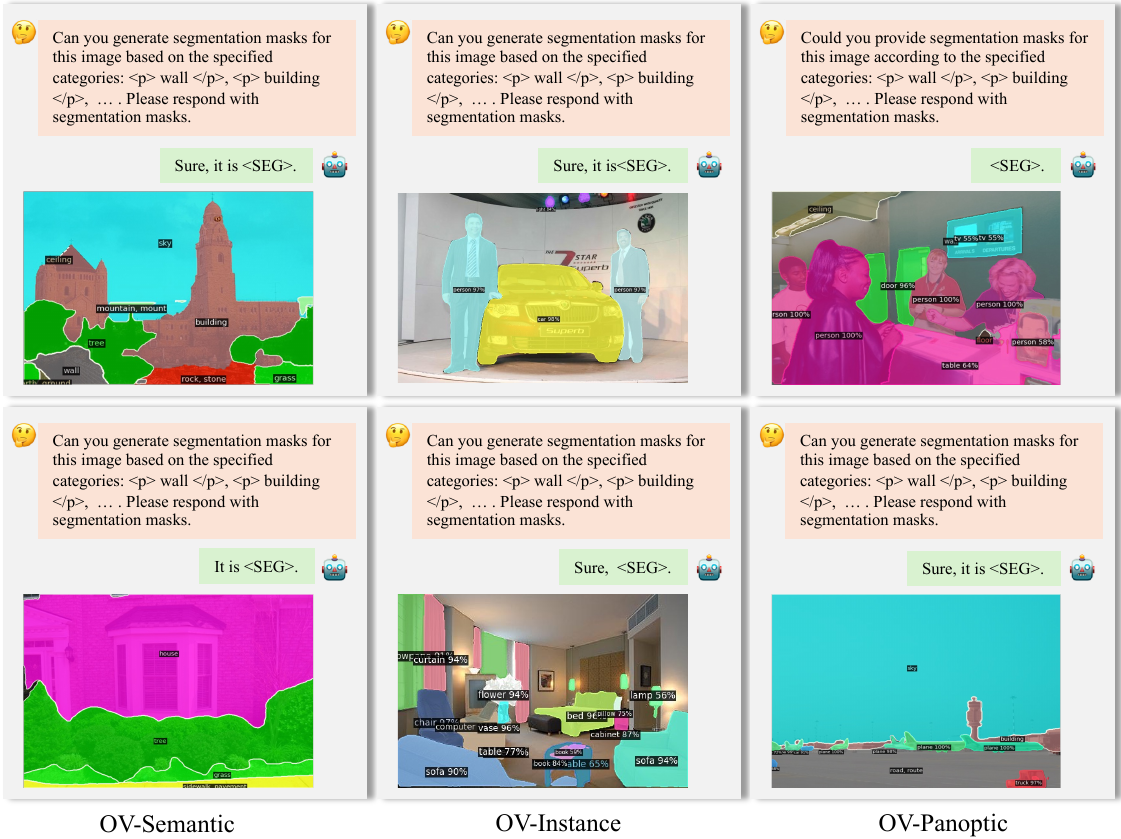}
    \captionsetup{skip=0pt,belowskip=-3pt}
    \caption{Visualization Results of Open-Vocabulary (OV) Segmentation. Visualized images are sampled from the ADE20K Val set. More category names are omitted for better visualization.}
    \label{fig:ovseg_vis}
\end{figure*}

\noindent\textbf{Open-Vocabulary Segmentation.} \figurename~\ref{fig:ovseg_vis} shows the visualization results of X-SAM in open-vocabulary (OV) segmentation, including OV-semantic, OV-instance, and OV-panoptic segmentation, which require segmenting objects that may not exist in the training set. X-SAM can segment objects that are not in the training set, demonstrating the robust generalization ability of the proposed method.

\begin{figure*}[ht]
    \centering
    \includegraphics[width=0.82\linewidth]{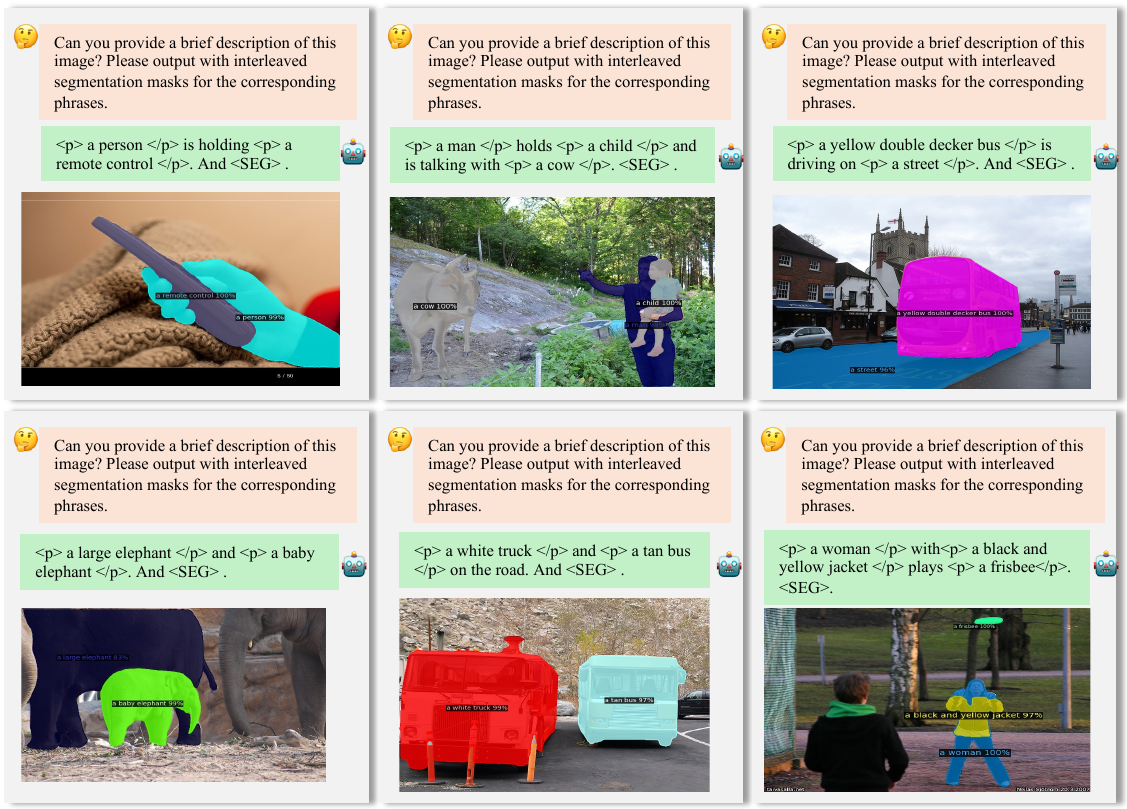}
    \captionsetup{skip=5pt,belowskip=-10pt}
    \caption{Visualization Results of GCG Segmentation. Visualized images are sampled from the Open-PSG Val set.}
    \label{fig:gcgseg_vis}
\end{figure*}

\noindent\textbf{GCG Segmentation.} \figurename~\ref{fig:gcgseg_vis} shows the visualization results of X-SAM in GCG segmentation, which needs to describe the image and output the corresponding mask. X-SAM can not only effectively understand the image and generate the language description, but also generate the segmentation masks for the corresponding phrases.

\begin{figure*}[ht]
    \centering
    \includegraphics[width=0.82\linewidth]{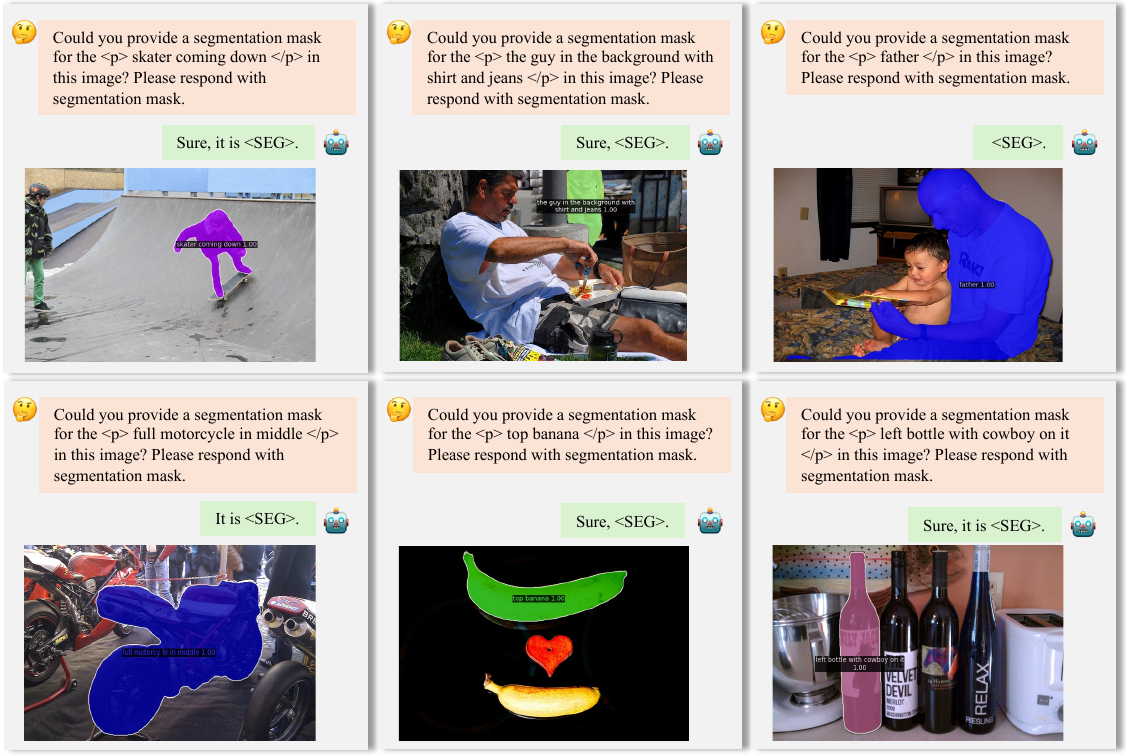}
    \caption{Visualization Results of Referring Segmentation. Visualized images are sampled from the RefCOCO Val set.}
    \label{fig:refseg_vis}
\end{figure*}

\noindent\textbf{Referring Segmentation.} \figurename~\ref{fig:refseg_vis} shows the visualization results of X-SAM in referring segmentation, which needs to segment objects that are referred to by natural language. X-SAM can effectively understand the referring expression and segment the objects that are referred to by natural language.

\begin{figure*}[ht]
    \centering
    \includegraphics[width=0.83\linewidth]{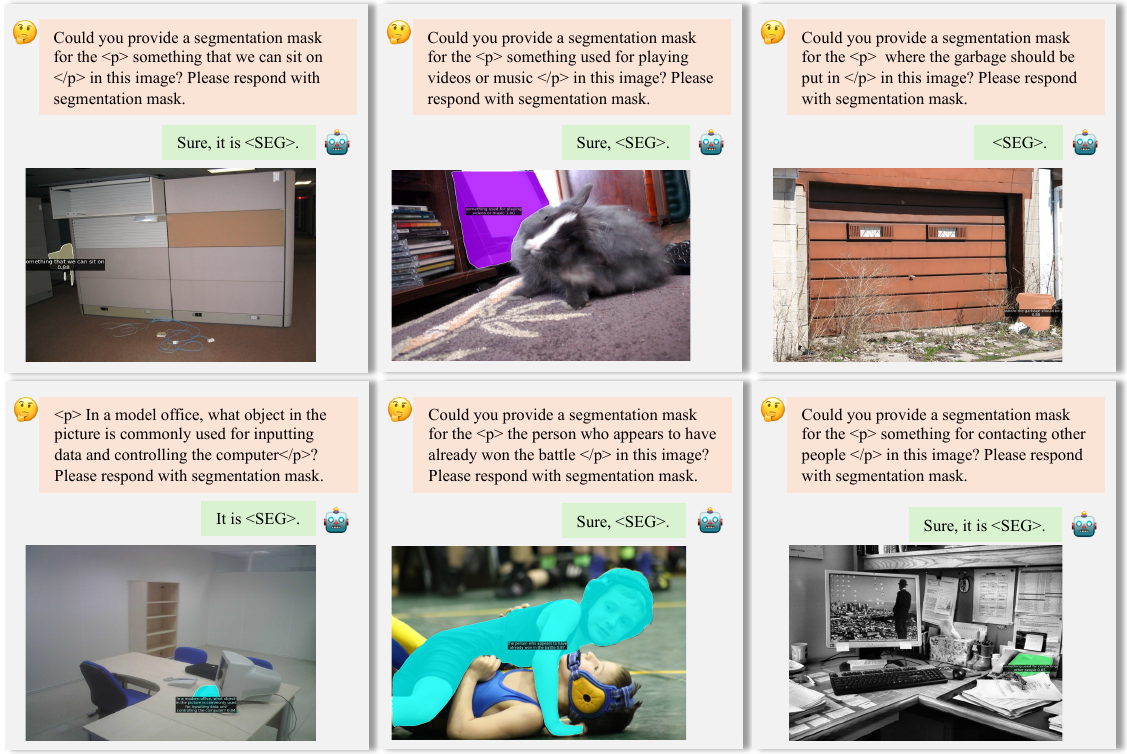}
    \captionsetup{skip=5pt,belowskip=-10pt}
    \caption{Visualization Results of Reasoning Segmentation. Visualized images are sampled from the reasoning segmentation Val set.}
    \label{fig:reaseg_vis}
\end{figure*}

\noindent\textbf{Reasoning Segmentation.} \figurename~\ref{fig:reaseg_vis} shows the visualization results of X-SAM in reasoning segmentation, which needs to segment the object that is related to the question. X-SAM can effectively understand the complex question and then generate the corresponding mask for the question.

\begin{figure*}[ht]
    \centering
    \includegraphics[width=0.95\linewidth]{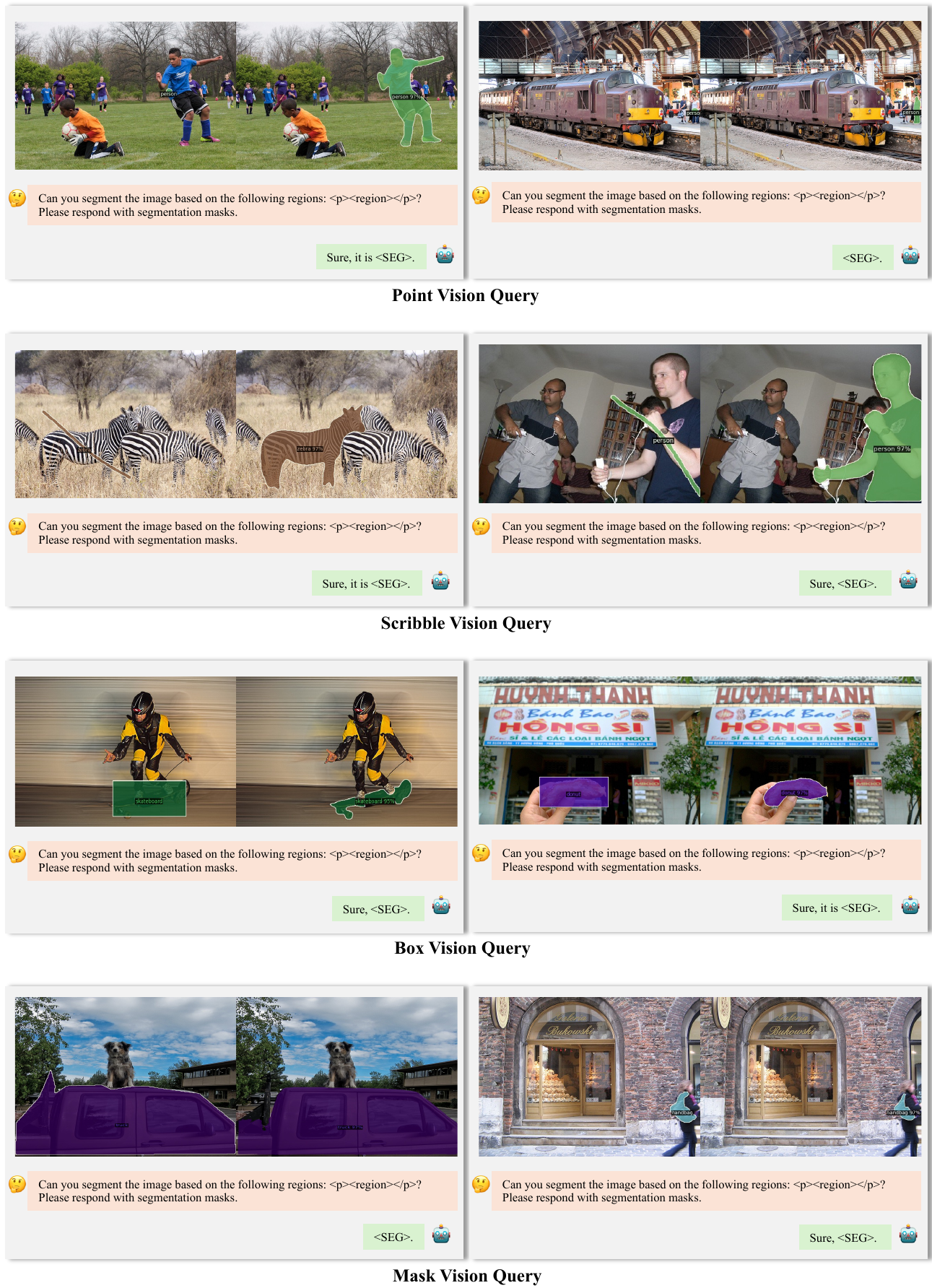}
    \caption{Visualization Results of Interactive Segmentation. Visualized images are sampled from the COCO-Interactive Val set. In each sample, the left image is the original image with the interactive prompts, and the right image is the visualized result of our method. From top to bottom, there are four types of visual interactive prompts: point, scribe, box, and mask. We visualize the category name of the interactive prompt for better visualization.}
    \label{fig:interseg_vis}
\end{figure*}

\noindent\textbf{Interactive Segmentation.} \figurename~\ref{fig:interseg_vis} shows the visualization results of X-SAM in interactive segmentation, which needs to segment the individual objects with which the user interacts. X-SAM can generate the corresponding mask for the user's interactive visual prompt.

\begin{figure*}[ht]
    \centering
    \includegraphics[width=0.95\linewidth]{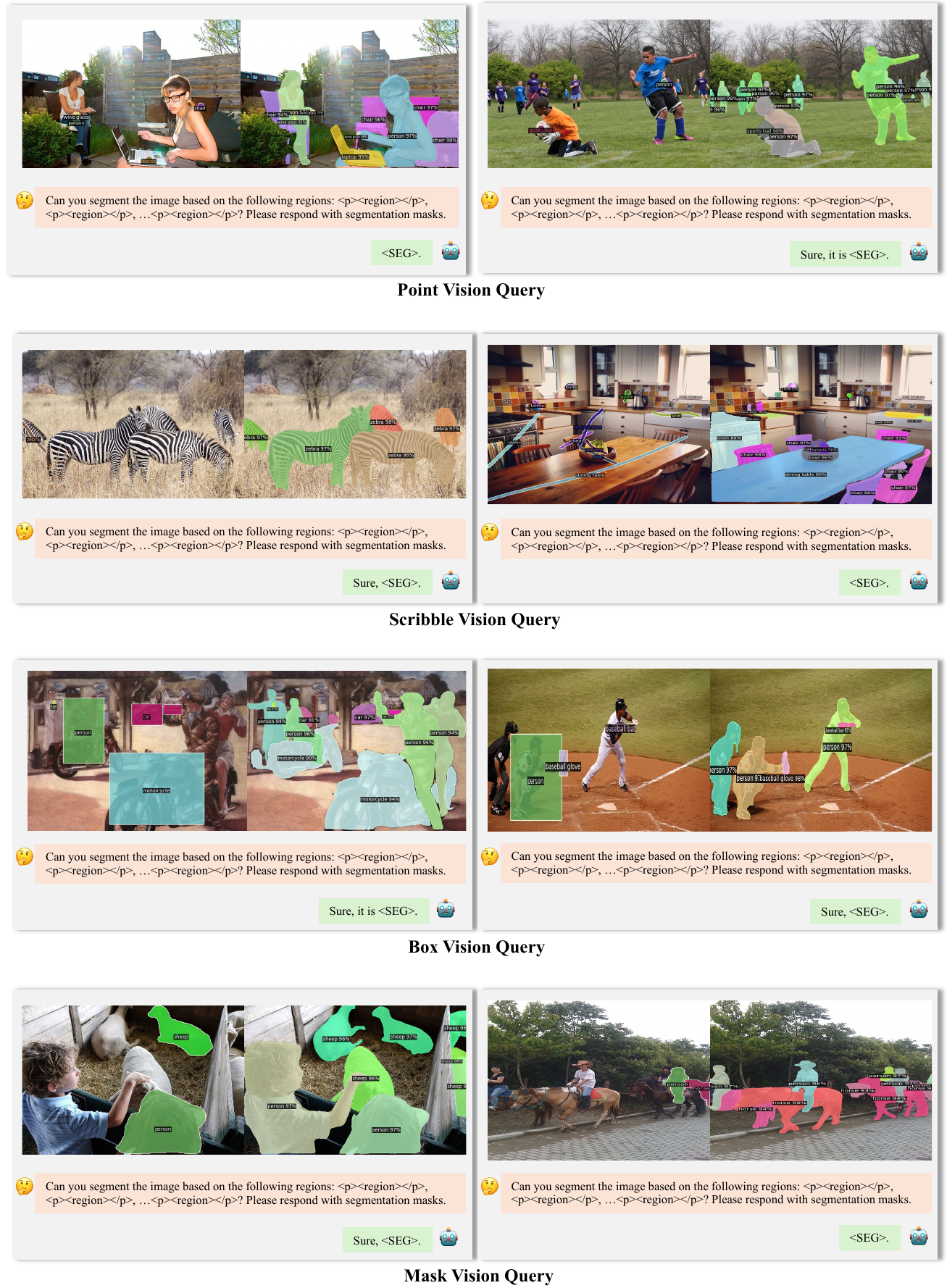}
    \caption{Visualization Results of VGD Segmentation (Single Image). Visualized images are sampled from COCO-VGD Val set. In each sample, the left image is the original image with the visual grounded prompts, and the right image is the visualized result of our method. From top to bottom, there are four types of visual grounded prompts: point, scribe, box, and mask. We visualize the category name of the grounded prompt for better visualization.}
    \label{fig:vgdseg_vis0}
\end{figure*}

\begin{figure*}[ht]
    \centering
    \includegraphics[width=0.95\linewidth]{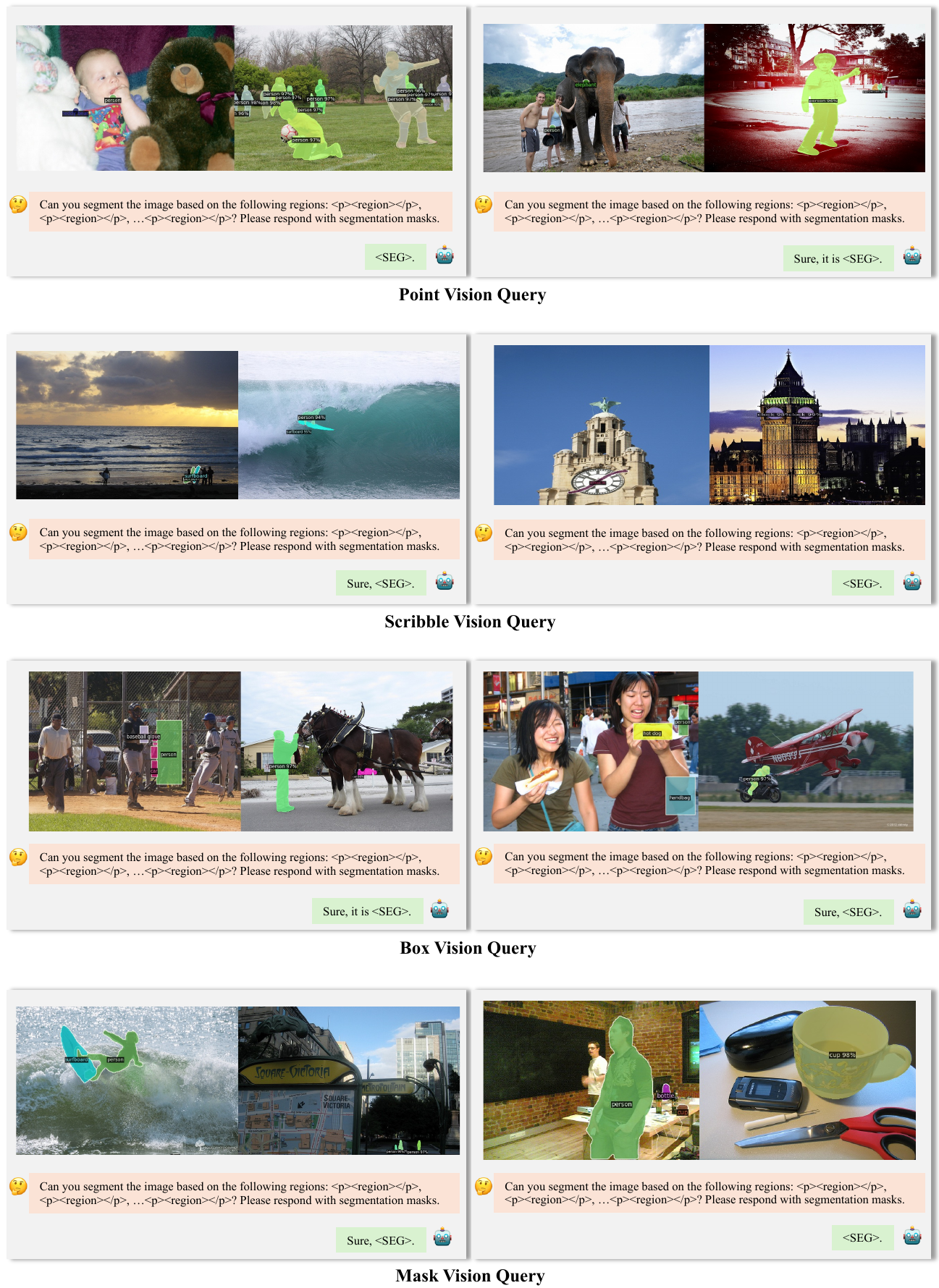}
    \caption{Visualization Results of VGD Segmentation (Cross Image). Visualized images are sampled from the COCO-VGD Val set. In each sample, the left image is a different image with the visual grounded prompts, and the right image is the visualized result of our method. From top to bottom, there are four types of visual grounded prompts: point, scribe, box, and mask. We visualize the category name of the grounded prompt for better visualization.}
    \label{fig:vgdseg_vis1}
\end{figure*}

\noindent\textbf{VGD Segmentation.} \figurename~\ref{fig:vgdseg_vis0} shows the visualization results of X-SAM in VGD segmentation for a single image, which needs to segment all the objects in the \textit{single image} that is grounded with the user's visual prompt. X-SAM can effectively segment all the objects in the image given by the user's visual grounded prompt. In addition, X-SAM can perform VGD segmentation on the \textit{cross image}, which needs to segment the objects grounded in another image. \figurename~\ref{fig:vgdseg_vis1} shows the visualization results of X-SAM in VGD segmentation for cross image, which demonstrates the effectiveness of X-SAM for VGD segmentation both in single image and cross image.

\subsection{Further Discussions}
\noindent\textbf{Limitation.} Although X-SAM achieves unified segmentation by extending \textit{segment anything} to \textit{any segmentation}, there is still considerable room for improvement. First, joint co-training with segmentation datasets and conversation datasets negatively impacts performance on some segmentation datasets, a phenomenon also observed in~\cite{zhang2024omgllava} and~\cite{rasheed2024glamm}. This challenge may be addressed by designing a more balanced dataset mixture. Second, the performance of X-SAM is not optimal across all tasks, which has also been observed in other unified segmentation methods~\cite{zhang2024omgllava,rasheed2024glamm,lai2024lisa}. This challenge remains a major obstacle for unified models and may be addressed by scaling up the model size and the amount of training data.

\noindent\textbf{Future Work.} Several avenues for future work can be explored with our novel unified framework. We highlight two potential directions. The first is to integrate X-SAM with SAM2~\cite{ravi2024sam2}, a unified model for segmentation in images and videos. This integration would further extend the application of X-SAM to video segmentation. The second direction is to extend VGD segmentation to the video domain, which would constitute an interesting video segmentation task and introduce visually grounded temporal information to segmentation. We plan to explore these directions in the future, provided that more computational resources become available.

\newpage
\end{document}